\documentclass{article}

\PassOptionsToPackage{numbers, compress}{natbib}

\usepackage[preprint]{neurips_2026} 

\usepackage[utf8]{inputenc} 
\usepackage[T1]{fontenc}    
\usepackage{hyperref}       
\usepackage{url}            
\usepackage{booktabs}       
\usepackage{caption}        
\usepackage{array}          
\usepackage{multirow}       
\usepackage{amsfonts}       
\usepackage{nicefrac}       
\usepackage{microtype}      
\usepackage{xcolor}         
\usepackage{enumitem}
\usepackage{amsmath}
\usepackage{amssymb}

\usepackage{kotex}
\usepackage{graphicx}
\usepackage{tabularx}
\usepackage{threeparttable}

\title{Beyond Attack Success Rate: Temporal Logit Observability for LLM Safety Failures}

\author{%
  Junyoung Park \\
  Chung-Ang University\\
  \texttt{june295921@cau.ac.kr} \\
  \and
  \textbf{Sunghwan Park} \\
  Chung-Ang University\\
  \texttt{tjdghks994@cau.ac.kr} \\
  \and
  \textbf{Seongyong Ju} \\
  Chung-Ang University\\
  \texttt{jusy4901@cau.ac.kr} \\
  \and
  \textbf{Jaewoo Lee}\thanks{Corresponding authors} \\
  Chung-Ang University\\
  \texttt{jaewoolee@cau.ac.kr} \\
}

\begin{document}
\maketitle

\begin{abstract}
Attack Success Rate (ASR) evaluates each jailbreak with a single yes/no label at the end of generation, telling us \emph{whether} a failure happened but not \emph{how} it unfolded.
Two attacks that produce equally harmful outputs may have followed completely different paths, and ASR cannot tell them apart.
We make those hidden paths observable from logits alone.
\emph{Temporal Logit Observability} (TLO) is a training-free diagnostic that watches a compliance--refusal margin during decoding and places each model--attack condition on a calibrated 2D plane.
By design, this plane is most informative exactly where ASR is least informative: among attacks that succeed for genuinely different reasons.
Across four aligned LLMs and three jailbreak paradigms, attacks with nearly identical ASR land at clearly different points on the plane: the same model can fail through different temporal patterns.
The geometry matches refusal-direction probes from hidden states on most conditions, with one model showing the limit of our fixed-lexicon approach.
A simple early-stop rule derived from TLO cuts successful jailbreaks by more than half, without false alarms on plain benign queries.
Safety evaluation should report \emph{when} and \emph{how} a failure unfolds, not only \emph{whether} it occurred. TLO makes the first two observable from logits alone. 
\end{abstract}

\section{Introduction}

Attack Success Rate (ASR) evaluates aligned large language models: a binary judgment of whether the final response refused or complied with a harmful prompt~\citep{mazeika2024harmbench, liu2023jailbreaking}.
ASR is operationally appealing and has become the de facto standard for jailbreak evaluation~\citep{christiano2017deep, ouyang2022training, bai2022helpful, bai2022constitutional}.
But ASR is a single number computed at the \emph{end} of generation: it treats decoding as a black box, recording the final label but not the decoding trajectory.

\begin{figure}[t]
    \centering
    \includegraphics[width=0.95\textwidth]{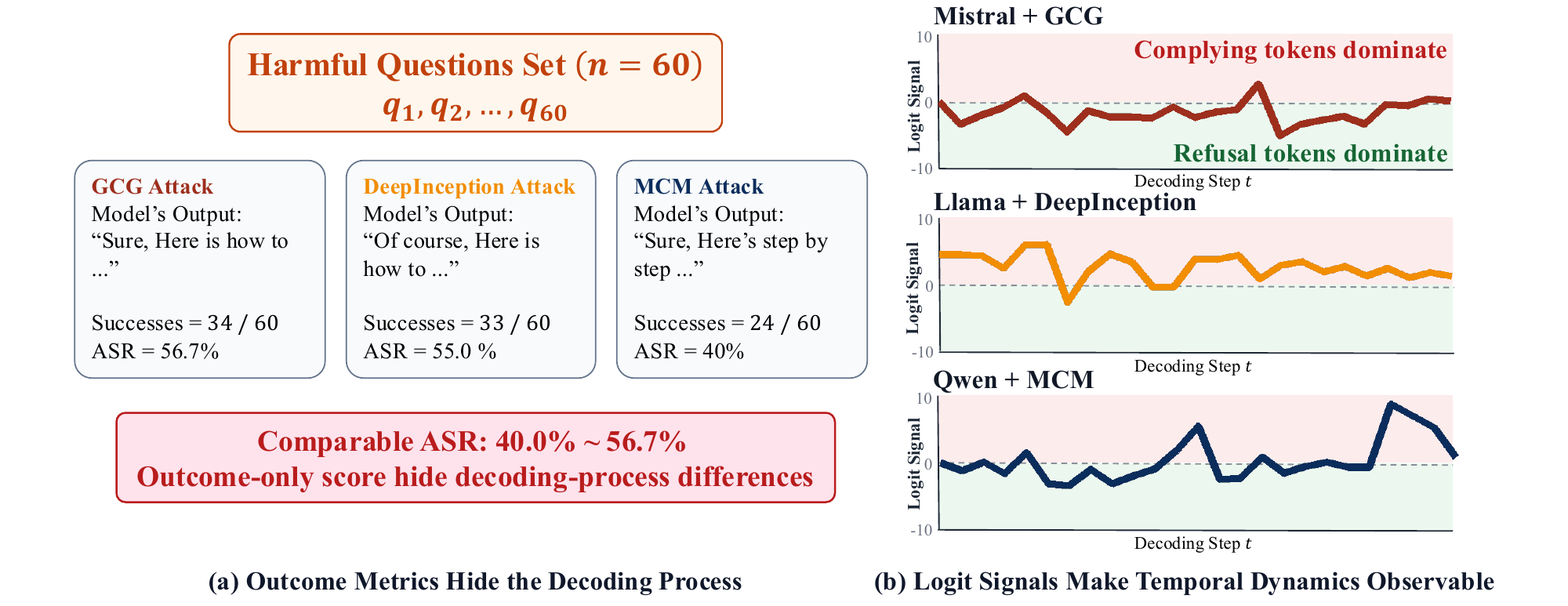}
    \caption{\textbf{ASR collapses temporally distinct safety failures.}
    \textbf{(a)}~Distinct attacks against the same harmful question can yield equally compliant final responses, indistinguishable under ASR.
    \textbf{(b)}~The model--attack conditions follow temporally distinct logit trajectories, recoverable as TLO's per-step compliance--refusal margin.}
    \label{fig:method_overview}
\end{figure}

This gap is a structural property of the metric: ASR by construction observes only the final response, discarding the trajectory by which it was produced.
As shown in Figure~\ref{fig:method_overview}(a), distinct attacks against the same harmful question can produce nearly indistinguishable final responses that ASR labels identically.
The same ASR-success label can hide three fundamentally different processes: refusal evidence may have appeared briefly and been suppressed; refusal may have emerged but too late to matter; or refusal may never have been triggered at all~\citep{ganguli2022red}.
Because ASR collapses these into one number, model--attack pairs with identical ASR may require different defenses that current evaluation cannot distinguish.

Existing approaches that recover this hidden structure rely on hidden-state probes~\citep{arditi2024refusal, zou2023representation, yin2025refusal}, which require activation access and model-specific instrumentation.
Output logits are far more widely available, yet existing logit-level safety work has focused on intervention, such as steering or decoding modification~\citep{li2025logit}, rather than diagnosis.
This leaves an open question: \emph{\textbf{can the temporal structure of safety failures hidden by ASR be recovered from logits alone, without hidden-state access?}}

\paragraph{Our approach.}
We introduce \emph{Temporal Logit Observability} (TLO), a training-free, logit-only diagnostic protocol that converts decoding-time trajectories into a calibrated two-dimensional coordinate.
TLO measures a compliance--refusal margin (the Logit-Margin Score, LMS) at each decoding step, summarizes it along pre-generation and generation-time axes, and calibrates each summary against harmful and attack-formatted benign references via Relative Position (RP).
The resulting RP-plane coordinate $c=(\mathrm{RP}_A, \mathrm{RP}_B)$ is comparable across models, attacks, and prompt formats, and exposes \emph{when} safety evidence concentrates without inspecting any internal state.
TLO complements ASR rather than replacing it.
By construction, RP-plane displacement compresses with $(1{-}\mathrm{ASR})$:
RP carries the primary signal in mid-ASR conditions, where ASR alone is least informative about \emph{why} attacks succeed, while ASR carries the load when failures are rare.

\paragraph{What TLO recovers.}
Across four aligned LLMs (Llama, Mistral, Qwen, Gemma) and three jailbreak paradigms (MCM, GCG, DI), TLO recovers structure that ASR systematically discards (Figure~\ref{fig:method_overview}(b)).
First, attacks with nearly identical ASR occupy clearly distinct locations in the RP plane: within attack DI, Llama and Qwen separate at $D_{\mathrm{RP}}=0.572$ despite an ASR difference of only 3.3\%, and across attack family the separation reaches $D_{\mathrm{RP}}=1.014$ at a 1.7\% ASR difference; the same attack family (GCG) further ranges from the RP origin on Qwen to a large generation-time displacement on Llama (\S\ref{sec:type_ab}).
Second, the underlying logit trajectory separates successful from failed jailbreaks within the first few generated tokens, and aligns with hidden-state refusal directions on 10 of 12 model--attack conditions, with model-specific scope clearly delimited (\S\ref{sec:signal_validation}).
Third, the geometry supports decoding-time intervention: a simple $t_{\text{cross}}$-derived early-stop rule reduces aggregate ASR from 39.6\% to 13.1\% with 0.0\% false-positive rate on format-free benign queries (\S\ref{sec:application}).
These observations are stable across decoding randomness, lexicon perturbations, judge variation, and stricter activation-timing rules (\S\ref{sec:robustness}).

\paragraph{Contributions.}
\begin{enumerate}[leftmargin=*, itemsep=1pt, parsep=0pt]
    \item \textbf{Calibrated temporal coordinates from logits alone.} The RP-plane coordinate $c=(\mathrm{RP}_A, \mathrm{RP}_B)$ summarizes the LMS trajectory along pre-generation and generation-time axes, calibrated against harmful and attack-formatted benign references. The $(1{-}\mathrm{ASR})$ bound on RP displacement is a deliberate complementarity with ASR (\S\ref{sec:method}).
    \item \textbf{Model--attack interactions invisible under ASR.}     Across 12 conditions, the same attack family produces qualitatively different temporal signatures across models, and the same model handles different attacks via different temporal mechanisms (\S\ref{sec:type_ab}). The recovered geometry is supported by hidden-state refusal directions on 10 of 12 conditions, with model-specific scope made explicit (\S\ref{sec:signal_validation}).
    \item \textbf{Decoding-time intervention.} A simple $t_{\text{cross}}$-derived halt rule reduces aggregate ASR from 39.6\% to 13.1\% with 0.0\% false alarms on format-free benign queries, demonstrating that the calibrated geometry drives effective decoding-time monitoring (\S\ref{sec:application}).
\end{enumerate}

\section{Preliminaries}
\label{sec:prelim}

We review ASR as the outcome-level metric that motivates TLO, then introduce the logit-only decoding notation used throughout the paper.

\paragraph{Evaluation at the final response.}
Attack Success Rate (ASR) formalizes jailbreak evaluation as a final-response outcome metric~\citep{mazeika2024harmbench, zou2023universal, liu2023jailbreaking}.
By aggregating final-response labels, ASR records whether an attack succeeded but discards the token-level trajectory by which the model moved toward refusal or compliance.

\paragraph{Decoding setup and access.}
We operate in a logit-only access setting.
Consider an autoregressive language model~\citep{vaswani2017attention, brown2020language} generating response tokens $y_1, \ldots, y_T$ given context $\mathbf{x} = (x_1, \ldots, x_n)$.
We compute one prompt-only diagnostic logit vector $\ell_0 \in \mathbb{R}^{|\mathcal{V}|}$ at the final prompt position before any response token is generated.
During generation, $\ell_t$ for $t=1, \ldots, T$ denotes the score vector returned immediately before sampling response token $y_t$.
The decoding index thus has one pre-generation position and $T$ generation-step positions; we write the observed score trajectory as $S_{0:T}$ after defining LMS below.
TLO uses only token-level logits to observe when safety-related evidence appears during decoding, without gradients, hidden states, architectural changes, or decoding-time intervention.
It requires full logits or targeted lexicon-token log-probabilities; standard top-$k$ outputs are insufficient (Appendix~\ref{appendix:topk_truncation}).
Hidden-state analyses serve as activation-based baselines for validation (\S\ref{sec:construct}).

\paragraph{Logit-Margin Score.}
We use the Logit-Margin Score (LMS) to measure the compliance--refusal balance at each decoding step.
\citet{li2025logit} introduce this margin as a \emph{steering} signal; we use the same observable for passive temporal diagnosis, and the calibration and analysis built on top of it form our contribution.
We aggregate compliance and refusal lexicons $\mathcal{L}_{\text{cmp}}, \mathcal{L}_{\text{ref}}$ (Appendix~\ref{app:lexicon}) into mean scores $\mu_{\text{cmp}}(t)$ and $\mu_{\text{ref}}(t)$, and define LMS as:
\begin{equation}
    S_t = \mu_{\text{cmp}}(t) - \mu_{\text{ref}}(t).
    \label{eq:lms}
\end{equation}
Positive $S_t$ indicates compliance-dominant logits; negative $S_t$ indicates refusal-dominant logits.
We use lexicon-internal top-$k$ aggregation with $k=10$; sensitivity is reported in Appendix~\ref{app:topk_aggregation}.

\section{Temporal Logit Observability}
\label{sec:method}

\begin{figure}[t]
    \centering
    \includegraphics[width=0.9\textwidth]{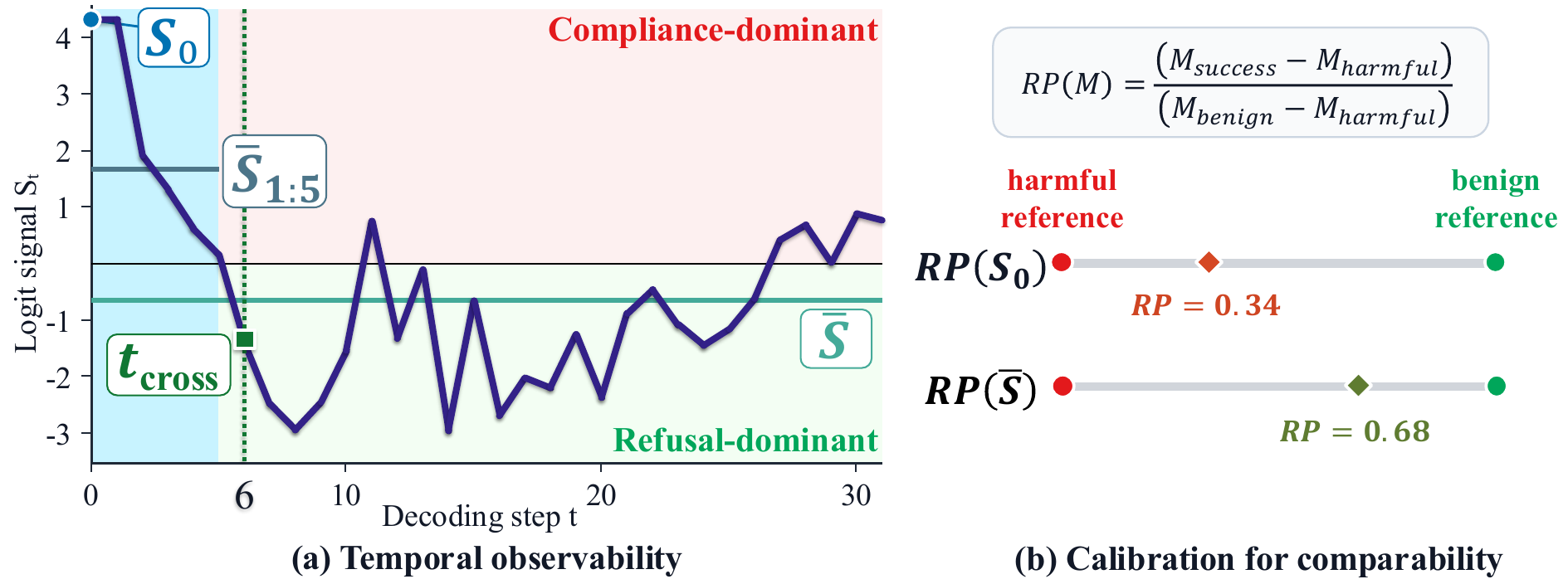}
    \caption{\textbf{Operationalizing TLO.}
    \textbf{(a)}~A real harmful-sample logit trajectory illustrates the external time series of compliance--refusal balance during generation.
    \textbf{(b)}~Relative-position calibration expresses each temporal observation relative to harmful and attack-formatted benign references, making conditions comparable without hidden-state access.}
    \label{fig:metrics_visual}
\end{figure}

Given an LMS trajectory $S_{0:T}$, the \emph{Temporal Logit Observability} (TLO) protocol converts decoding-time logit observations into calibrated 2D coordinates without hidden-state access or model modification.
TLO and ASR are complementary by design: TLO is most informative when attacks often succeed (where ASR alone says little about \emph{why}), while ASR is most informative when attacks rarely succeed.
TLO proceeds in three stages: (i) a per-step temporal logit signal (\S\ref{sec:signal_axes}); (ii) activation-timing summaries such as crossing time and sign reversal (\S\ref{sec:activation_timing}); and (iii) calibration against harmful and attack-formatted benign references via Relative Position (\S\ref{sec:rp_classification}).
Figure~\ref{fig:metrics_visual} summarizes the pipeline.

\subsection{Temporal Logit Signal}
\label{sec:signal_axes}

TLO computes the Logit-Margin Score $S_t$ from Eq.~\ref{eq:lms} at each decoding step, yielding a trajectory $S_{0:T}$.
From this trajectory, we record three compliance--refusal summaries that drive activation timing and RP calibration.

$S_0$ is the \emph{pre-generation observation}, computed at the final prompt position before any response token is generated.
It captures whether the input context has already shifted the model's logit state toward compliance before decoding begins.

$\bar{S} = \frac{1}{T}\sum_{t=1}^{T} S_t$ is the \emph{generation-time observation}, computed as the mean logit signal over the $T$ generation steps.
A prefix variant $\bar{S}_{1:w} = \frac{1}{w}\sum_{t=1}^{w} S_t$ provides an early-window summary.
$S_0$ and $\bar{S}$ jointly form the axes of the RP-calibrated plane in \S\ref{sec:rp_classification}.

\subsection{Activation Timing Signals}
\label{sec:activation_timing}

Activation-timing signals measure when the LMS trajectory first shows refusal-dominant evidence.

$t_{\text{cross}} = \min\{t \geq 1 : S_t < 0\}$ records the first generation step at which refusal dominates ($t \geq 1$ excludes the pre-generation position $S_0$).
If no crossing occurs within the generated sequence, $t_{\text{cross}}$ is right-censored.
For ROC analysis, censored samples are tied at the maximum crossing rank within the analysis horizon, not ordered by response length.
A stricter $n$-consecutive-step variant is verified in Appendix~\ref{app:tcross_sensitivity}.

\emph{Sign reversal} ($S_0 > 0 \land \bar{S} < 0$) records a sustained shift from an initially compliance-dominant trajectory to refusal dominance during generation.
We compute sign reversal only on harmful prompts, as benign sign patterns are format-dependent rather than safety-driven (Appendix~\ref{app:benign_exclusion}).

\subsection{Relative-Position Calibration}
\label{sec:rp_classification}

Raw logit observations cannot be directly compared across models, attacks, or prompt formats: a positive $S_t$ on Qwen and on Llama do not mean the same thing.
We therefore calibrate each observation through \textbf{Relative Position (RP)}, which expresses successful attacks relative to harmful and attack-formatted benign distributions:
\begin{equation}
    \text{RP}(M) = \frac{M_{\text{success}} - M_{\text{harmful}}}{M_{\text{benign}} - M_{\text{harmful}}},
    \label{eq:rp}
\end{equation}
where $M_{\text{success}}$, $M_{\text{harmful}}$, and $M_{\text{benign}}$ are condition means over successful attacks, all harmful samples, and attack-formatted benign references.
$\text{RP}=0$ indicates harmful-like behavior; $\text{RP}=1$ indicates benign-like behavior; values outside $[0,1]$ are reported as observed inversions rather than clipped.
The all-harmful and attack-formatted benign anchors avoid any need for failed-jailbreak samples while controlling for the attack wrapper.
Anchor sensitivity, denominator audits, and a rank-based variant are reported in Appendix~\ref{app:rp_sensitivity_checks}.

Applying Eq.~\ref{eq:rp} to $S_0$ yields $\mathrm{RP}_A$; applying it to $\bar{S}$ yields $\mathrm{RP}_B$.
We treat the resulting two-dimensional coordinate $c=(\mathrm{RP}_A, \mathrm{RP}_B)$ as the primary calibrated observation, and define RP-plane divergence between conditions $a$ and $b$ as the Euclidean distance:
\begin{equation}
    D_{\mathrm{RP}}(a,b) = \|c_a - c_b\|_2.
    \label{eq:rp_distance}
\end{equation}
$D_{\mathrm{RP}}$ is measured in RP-normalized units: each axis is scaled by the harmful-to-benign reference gap of its temporal summary, so one unit equals the harmful--benign anchor distance for that summary.
Higher $D_{\mathrm{RP}}$ indicates greater calibrated temporal separation, even when final ASR is similar.

\paragraph{What RP measures, by design.}
Because the harmful anchor pools all harmful samples, $M_{\mathrm{success}} - M_{\mathrm{harmful}} = (1-\mathrm{ASR})(M_{\mathrm{success}} - M_{\mathrm{failed}})$, so RP-plane displacement is bounded by $(1-\mathrm{ASR})$ on each axis.
This formalizes the complementarity introduced at the start of \S\ref{sec:method}: as ASR rises, RP shrinks toward zero by design.
High-ASR conditions therefore carry their primary signal in ASR itself, while mid-ASR conditions are where RP separates attacks that ASR cannot.

\paragraph{Reading the RP plane.}
The continuous coordinate $c$ is the primary statistic.
For readability, we attach significance-filtered axis annotations: a label-permutation test on the displacement $r=\|c\|_2$ with Benjamini--Hochberg FDR at $q=0.05$ identifies conditions with reliably non-zero displacement (Appendix~\ref{app:permutation_test}).
For displaced conditions, we mark the dominant axis as \textbf{Pre-generation Bias} (PGB) when $|\mathrm{RP}_A|>|\mathrm{RP}_B|$, and \textbf{In-generation Bias} (IGB) when $|\mathrm{RP}_B|>|\mathrm{RP}_A|$.
Conditions failing the permutation test are annotated as \textbf{Null-generation Bias} (NGB), meaning no reliable RP-axis displacement under the calibrated protocol.
These labels mark axis dominance throughout the paper; axis-margin and small-success caveats are reported in Appendix~\ref{app:axis_stability}.

\section{Experiments}
\label{sec:experiments}

The experiments answer three questions about TLO.
\textbf{(Q1)} Does the calibrated RP plane recover the model--attack structure that ASR collapses (\S\ref{sec:type_ab})?
\textbf{(Q2)} Are the underlying logit signals informative and aligned with hidden-state safety representations (\S\ref{sec:signal_validation})?
\textbf{(Q3)} Can the recovered geometry drive action during decoding (\S\ref{sec:application})?
We then summarize robustness checks (\S\ref{sec:robustness}).

\begin{figure}[!tbp]
    \centering
    \includegraphics[width=0.78\textwidth]{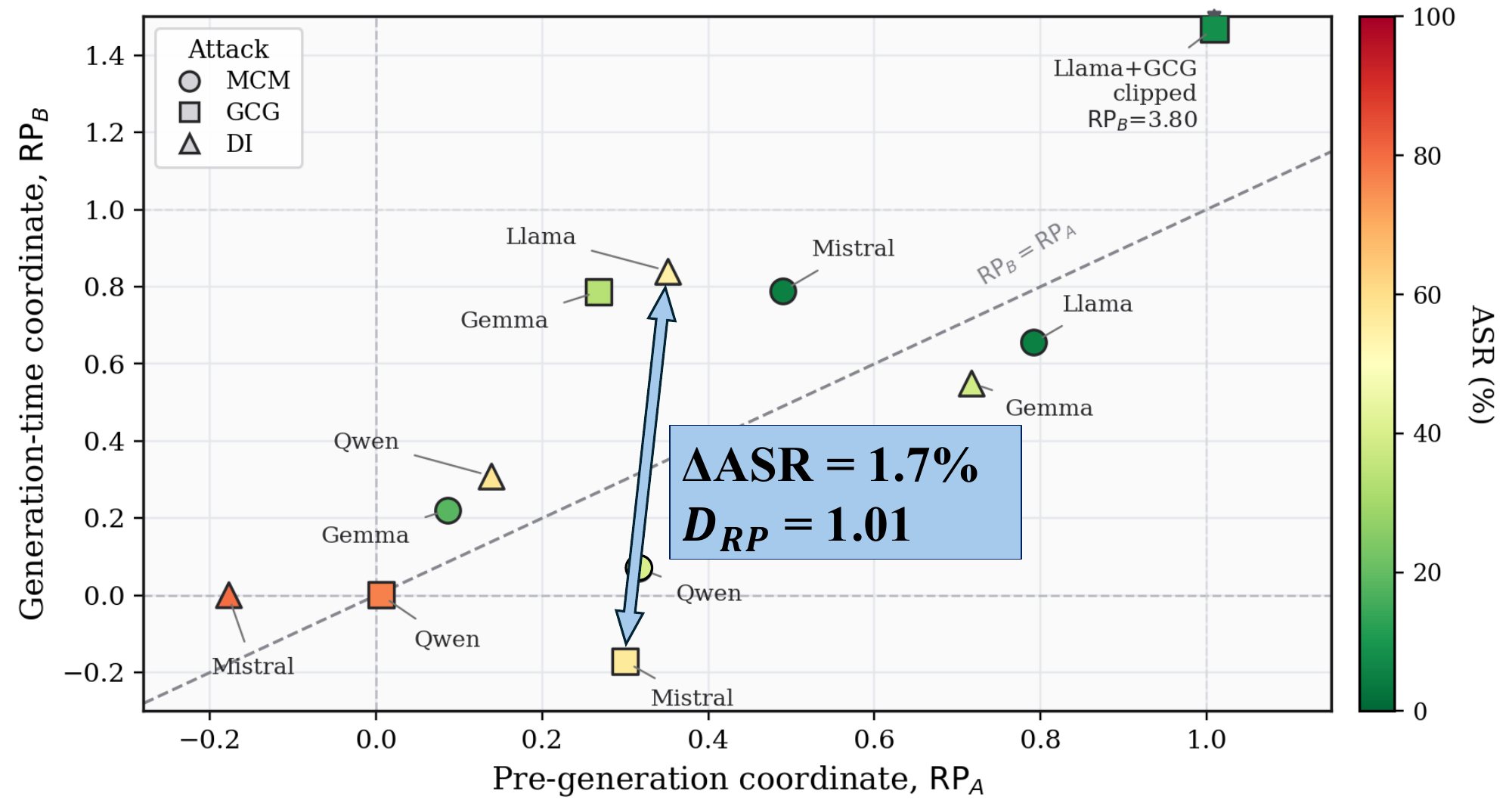}
    \caption{\textbf{Conditions with similar ASR can occupy distinct RP-plane locations.}
    Each point is a model--attack condition with coordinate $c=(\mathrm{RP}_A,\mathrm{RP}_B)$ that captures pre-generation and generation-time calibrated logit position; marker shape denotes attack family and color denotes ASR under the primary Llama-Guard judge.
    The diagonal marks $\mathrm{RP}_B=\mathrm{RP}_A$ for visual reference.
    Llama+GCG has $\mathrm{RP}_B=3.80$ and is clipped for readability; all analyses use unclipped coordinates.}
    \label{fig:rp_plane_asr_spectrum}
\end{figure}

\subsection{Setup}
\label{sec:setup_exp}

We evaluate TLO on a 4$\times$3 grid of four open instruct models (Llama~\citep{grattafiori2024llama3}, Mistral~\citep{jiang2023mistral}, Qwen~\citep{yang2024qwen25}, Gemma~\citep{gemma2team2024}) and three jailbreak paradigms, with 60 harmful JailbreakBench prompts~\citep{chao2024jailbreakbench} per condition under greedy decoding.
The attack grid covers multi-turn context manipulation (MCM; Appendix~\ref{app:mcm}), suffix optimization (GCG~\citep{zou2023universal}), and template-based semantic jailbreaks (DeepInception, DI~\citep{li2023deepinception}).
For RP calibration, each condition also includes 60 attack-formatted benign references that preserve the wrapper but replace the harmful request, yielding 720 harmful generations and 720 attack-formatted benign generations.
Attack success is judged by Llama-Guard-3-8B~\citep{inan2023llama}; auxiliary judges and a human audit are reported in Appendix~\ref{app:evaluation_grid}--\ref{app:judge}.

\subsection{Q1: RP-Plane Geometry Recovers ASR-Hidden Model--Attack Interactions}
\label{sec:type_ab}

Our central claim is that conditions with similar ASR can occupy distinct locations in the RP plane.
We use the continuous coordinate $c=(\mathrm{RP}_A, \mathrm{RP}_B)$ and pairwise $D_{\mathrm{RP}}$ as primary observations; PGB/IGB/NGB labels (Eq.~\ref{eq:rp_classify}) are descriptive axis annotations.
Figure~\ref{fig:rp_plane_asr_spectrum} visualizes the 12 condition coordinates with ASR as color, and Table~\ref{tab:type_ab} reports coordinates, displacement magnitude, permutation $p$-values, RP-plane annotations, and the early-stop probe response (the probe itself is reported in \S\ref{sec:application}).

\begin{table}[t]
    \centering
    \caption{\textbf{RP-plane coordinates separate model--attack conditions that ASR collapses.}
    Each row reports a condition's calibrated 2D position, displacement magnitude $r$, permutation significance, and the early-stop probe's intervention response (\S\ref{sec:application}).
    PGB = Pre-generation Bias; IGB = In-generation Bias; NGB = Null-generation Bias (high-ASR range where displacement shrinks by design).}
    \label{tab:type_ab}
    \scriptsize
    \resizebox{\textwidth}{!}{%
    \begin{tabular}{@{}llccccclccc@{}}
        \toprule
        & & \multicolumn{6}{c}{\textbf{RP-plane coordinates and annotation}} & \multicolumn{3}{c}{\textbf{Behavioral probe} ($w=5$)} \\
        \cmidrule(lr){3-8} \cmidrule(lr){9-11}
        Model & Attack & ASR & $\mathrm{RP}_A$ & $\mathrm{RP}_B$ & $r$ & $p_{\text{perm}}$ & Annotation & ASR$_{\text{orig}}$ & ASR$_{\text{probe}}$ & $\Delta$ASR \\
        \midrule
        \multirow{3}{*}{Llama}   & GCG           &  8.3\% & 1.01 & 3.80       & 3.94 & ${<}.001$ & \textbf{IGB} &  8.3\% &  1.7\% &  6.7 \\
                                  & MCM           &  5.0\% & 0.79 & 0.66       & 1.03 & ${<}.001$ & \textbf{PGB} &  5.0\% &  5.0\% &  0.0 \\
                                  & DI            & 55.0\% & 0.35 & 0.84       & 0.91 & ${<}.001$ & \textbf{IGB} & 55.0\% &  8.3\% & 46.7 \\
        \cmidrule(lr){1-11}
        \multirow{3}{*}{Mistral} & GCG           & 56.7\% & 0.30 & $-$0.17    & 0.35 & ${<}.001$ & \textbf{PGB} & 56.7\% & 53.3\% &  3.3 \\
                                  & MCM           &  5.0\% & 0.49 & 0.79       & 0.93 & $.056$    & \textit{NGB} &  5.0\% &  5.0\% &  0.0 \\
                                  & DI            & 80.0\% & $-$0.18 & 0.00   & 0.18 & $.506$    & \textit{NGB} & 80.0\% &  0.0\% & 80.0 \\
        \cmidrule(lr){1-11}
        \multirow{3}{*}{Qwen}    & MCM           & 40.0\% & 0.32 & 0.07       & 0.32 & $.271$    & \textit{NGB} & 40.0\% & 40.0\% &  0.0 \\
                                  & GCG           & 76.7\% & 0.01 & 0.00       & 0.01 & $.987$    & \textit{NGB} & 76.7\% &  8.3\% & 68.3 \\
                                  & DI            & 58.3\% & 0.14 & 0.31       & 0.34 & $.074$    & \textit{NGB} & 58.3\% &  5.0\% & 53.3 \\
        \cmidrule(lr){1-11}
        \multirow{3}{*}{Gemma}   & GCG           & 33.3\% & 0.27 & 0.79       & 0.83 & $.003$    & \textbf{IGB} & 33.3\% & 11.7\% & 21.7 \\
                                  & MCM           & 18.3\% & 0.09 & 0.22       & 0.24 & ${<}.001$ & \textbf{IGB} & 18.3\% & 18.3\% &  0.0 \\
                                  & DI            & 38.3\% & 0.72 & 0.55       & 0.90 & ${<}.001$ & \textbf{PGB} & 38.3\% &  0.0\% & 38.3 \\
        \midrule
        \multicolumn{8}{r}{\textbf{Aggregate}} & 39.6\% & 13.1\% & \textbf{26.5} \\
        \bottomrule
    \end{tabular}%
    }
\end{table}

\paragraph{ASR-neighboring conditions occupy distinct RP-plane locations.}
Among nine model--attack pairs with ASR within 5\%, the median calibrated separation is $D_{\mathrm{RP}}=0.572$ with stratified-bootstrap 95\% CI [0.489, 0.924] (Appendix~\ref{app:rp_sensitivity_checks}, Table~\ref{tab:drp_neighbor_ci}).
The cleanest within-attack pair, Llama+DI vs.\ Qwen+DI (ASRs 55.0\% vs.\ 58.3\%, $\Delta=$3.3\%), separates at $D_{\mathrm{RP}}=0.572$ [0.237, 1.026], showing that the same attack on different models produces distinct calibrated geometries even at near-equivalent ASR.
The strongest ASR-neighbor pair, Llama+DI vs.\ Mistral+GCG (ASRs 55.0\% vs.\ 56.7\%, $\Delta=$1.7\%), reaches $D_{\mathrm{RP}}=1.014$ [0.726, 1.351], adding attack identity to model identity as a second source of mechanism difference.
For reference, within-condition stratified-bootstrap intervals on the displacement magnitude have a median 95\% CI width of about 0.6 RP units (Appendix~\ref{app:permutation_test}, Table~\ref{tab:r_bootstrap}); both pair separations exceed this typical within-condition spread.
ASR-neighbor conditions therefore have nearly identical safety outcomes but qualitatively different decoding-time geometries: structure that ASR collapses by design.

\paragraph{Attack family does not determine RP-plane location.}
The same attack family occupies different RP-plane locations across models.
GCG ranges from Qwen near the RP origin $(0.01, 0.00)$ (expected from the $(1{-}\mathrm{ASR})$ bound at ASR $76.7\%$) to Llama's large generation-time displacement $(1.01, 3.80)$ at low ASR $8.3\%$, with Mistral and Gemma occupying intermediate regions (PGB and IGB respectively).
Llama+GCG's $\mathrm{RP}_B=3.80$ exceeds the $[0,1]$ reference interval because successful samples lie beyond the benign anchor on the generation-time axis; this inversion is stable under rank-based RP and bootstrap resampling (Appendix~\ref{app:rp_sensitivity_checks}, \ref{app:permutation_test}).
Defenses tuned to a single attack family but uniformly applied across models will therefore see condition-dependent effects that ASR cannot predict.

\paragraph{Model identity does not determine RP-plane location.}
The same model occupies different RP-plane locations under different attacks.
Llama places MCM $(0.79, 0.66)$ (PGB) apart from DI $(0.35, 0.84)$ (IGB), and Gemma places DI $(0.72, 0.55)$ (PGB) apart from MCM $(0.09, 0.22)$ (IGB).
A single per-model defense without attack-aware adaptation therefore mismatches the underlying temporal structure.

\paragraph{Reliable conditions span the grid.}
Of the 12 conditions, 7 show statistically reliable RP-plane displacement under FDR-corrected permutation, and these 7 span all 4 models and all 3 attacks.
The remaining 5 are either high-ASR (NGB by design; e.g., Mistral+DI at ASR $80\%$, Qwen+GCG at ASR $76.7\%$) or have low success counts that produce wide bootstrap intervals (Llama+MCM, Mistral+MCM, $n_{\text{succ}}=3$).
An independent entropy-based check confirms NGB reflects the high-ASR range rather than lexicon insensitivity (Appendix~\ref{app:entropy_crossval}).

\subsection{Q2: The Logit Signal Is Informative and Tracks Hidden-State Safety Representations}
\label{sec:signal_validation}

The RP geometry of \S\ref{sec:type_ab} rests on the LMS trajectory and on the assumption that this trajectory carries safety-relevant information.
We verify both: the trajectory predicts safety outcomes within the first few generated tokens, and aligns with hidden-state refusal directions on most conditions, with model-specific scope made explicit.

\begin{figure}[t]
    \centering
    \includegraphics[width=0.95\textwidth]{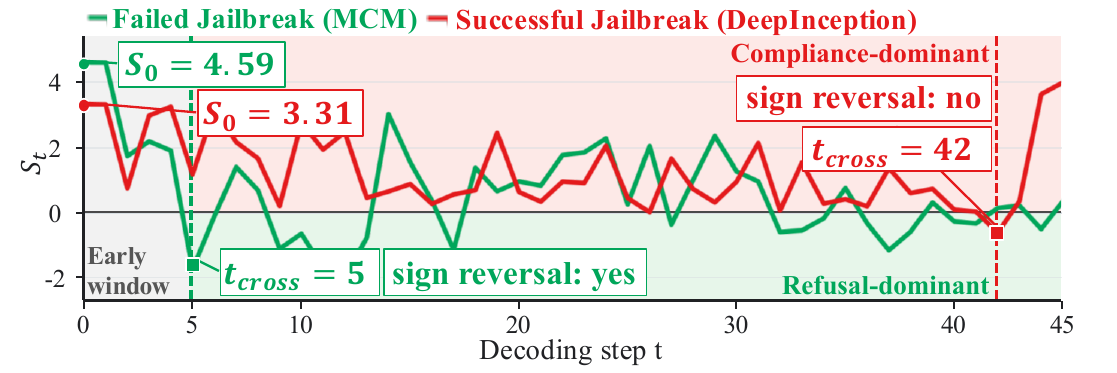}
    \caption{\textbf{TLO marks safety-activation timing on the trajectory.}
    For the same harmful question on Mistral under two attack wrappers, the failed MCM jailbreak crosses early ($t_{\text{cross}}=5$) and shows sign reversal, while the successful DI jailbreak delays crossing ($t_{\text{cross}}=42$) without sign reversal.}
    \label{fig:temporal_obs}
\end{figure}

\paragraph{The trajectory separates outcomes within the first five tokens.}
\label{sec:activation}
Failed jailbreaks typically cross into refusal dominance early, while successful jailbreaks delay or avoid that transition (Figure~\ref{fig:temporal_obs}).
Within the first five generated tokens, successful and failed jailbreaks separate at Cohen's $d > 1.0$ ($p < 10^{-38}$, $n=720$): the safety outcome is observable in the logit margin long before the response is complete.
$\bar{S}_{1{:}5}$ achieves ROC-AUC~\citep{fawcett2006introduction} $=0.771$ [0.737, 0.804] and $t_{\text{cross}}$ achieves $0.784$ [0.750, 0.816].
A trained 13-feature logistic regression on the same trajectory features reaches AUC $=0.826$ (Appendix~\ref{app:learned_baseline}), so the fixed-form summaries used here serve as an interpretable lower bound on extractable trajectory information.
Failed jailbreaks exhibit sign reversal $5.2{\times}$ more often than successful ones (40.0\% vs.\ 7.7\%), and successful jailbreaks remain compliance-dominant much more often (92.3\% vs.\ 49.7\%); delayed or absent refusal activation is therefore the logit-level signature TLO observes (token-rank evidence in Appendix~\ref{app:token_rank_linkage}).

\paragraph{The trajectory aligns with hidden-state refusal directions, with model-specific scope.}
\label{sec:construct}
We fix the hidden layer at $\sim$50\% network depth (L50), compute a per-model difference-of-means refusal direction from paired harmful/benign references without success labels, and project decoding-time hidden states.
The expected $S_t$--$z_t$ correlation $\rho$ is negative, summarized over steps 0--4 (full protocol, layer sweep, and probe-AUC comparisons in Appendix~\ref{app:hidden_crossmodel}).

Alignment strength varies by model.
On Llama, all three attacks reach $|\rho|\geq 0.70$, and the modal cross-correlation lag is $-1$: the logit margin moves \emph{one decoding step earlier} than the refusal-direction projection, consistent with autoregressive feedback in which the $t{-}1$ output token modulates $\mathbf{h}_t$ through the input context.
Mistral and Gemma preserve the expected sign with moderate strength (peaking at $|\rho|=0.46$ for Mistral+GCG and $|\rho|=0.50$ for Gemma+DI).
Qwen marks a model-specific boundary: it reverses the expected sign under MCM and DI, while only Qwen+GCG retains it.
The Fisher-$z$ mean across Qwen conditions is approximately zero, a known limit of the fixed-lexicon protocol on Qwen rather than a contradiction of the cross-model pattern.
Model-specific or probe-derived lexicons recover more evidence in such cases (Appendix~\ref{app:lexicon_induction}, \ref{app:probe_lexicon}).

Across the 12 conditions, 10 show the expected sign between $S_t$ and refusal-direction projections (Table~\ref{tab:hidden_temporal_alignment}); layer-sweep checks at L25, L50, and L75 confirm this is not a layer-tuned result (Appendix~\ref{app:hidden_layer_sweep}).

\begin{table}[h]
    \centering
    \caption{\textbf{Hidden-state alignment summary.}
    Expected-sign alignment between $S_t$ and fixed-L50 hidden-state projections across model--attack conditions.}
    \label{tab:hidden_temporal_alignment}
    \footnotesize
    \setlength{\tabcolsep}{3.0pt}
    \renewcommand{\arraystretch}{1.18}
    \begin{tabular}{@{}lcccccccccccc@{}}
        \toprule
        & \multicolumn{3}{c}{Llama} & \multicolumn{3}{c}{Qwen} & \multicolumn{3}{c}{Mistral} & \multicolumn{3}{c}{Gemma} \\
        \cmidrule(lr){2-4} \cmidrule(lr){5-7} \cmidrule(lr){8-10} \cmidrule(l){11-13}
        Hidden-state alignment & MCM & GCG & DI & MCM & GCG & DI & MCM & GCG & DI & MCM & GCG & DI \\
        \midrule
        Strong alignment & \checkmark & \checkmark & \checkmark & & & & & & & & & \\
        \cmidrule(lr){2-13}
        Aligned & & & & & \checkmark & & \checkmark & \checkmark & \checkmark & \checkmark & \checkmark & \checkmark \\
        \cmidrule(lr){2-13}
        Boundary & & & & \checkmark & & \checkmark & & & & & & \\
        \bottomrule
    \end{tabular}
\end{table}

\subsection{Q3: From Observation to Action via RP-Aware Early Stopping}
\label{sec:application}

The RP-plane geometry can drive action during decoding.
We derive a halt rule directly from $t_{\text{cross}}$: halt when $t_{\text{cross}}>5$ and replace the continuation with a standard refusal, using the fixed early-activation window of \S\ref{sec:activation}.

\paragraph{Aggregate effect.}
With $w=5$, aggregate ASR drops from 39.6\% to 13.1\% ($\Delta=26.5$\%), and the response varies systematically by RP-plane region (PGB $\Delta=$13.9\%, IGB $\Delta=$18.8\%, NGB $\Delta=$40.3\%; Appendix~\ref{app:intervention_judges}).
The largest drops occur in NGB conditions, consistent with the complementarity of \S\ref{sec:method}: where RP geometry is uninformative by design, ASR carries the primary signal and a uniform halt rule is most effective.

\paragraph{Benign behavior.}
Format-free benign queries had 0.0\% false positives across all 12 conditions.
Attack-formatted benign references, where benign prompts are deliberately wrapped in the attack template, share the same temporal cues used to detect attacks; an $S_0$-gated variant that requires a pre-generation harmful signal reduces false halts on these stress prompts from 46.0\% to 12.6\% (Appendix~\ref{app:s0_gating}).

\subsection{Robustness}
\label{sec:robustness}

The temporal and RP-plane structure persists across decoding randomness (ICC $=0.90$--$0.94$), stricter confirmed-crossing rules, a trained 13-feature classifier ($\Delta\text{AUC}=0.04$ over fixed $t_{\text{cross}}$), non-lexical controls (entropy, logit norm), lexicon perturbations ($|\Delta\text{AUC}|\leq 0.018$ across five variants), metric decomposition ($\sim$0.16 AUC margin gain over single-component variants), and data-driven lexicon induction (Appendix~\ref{app:stochastic}--\ref{app:lexicon_induction}).
TLO is therefore not a surface refusal-token detector or a compliance-onset heuristic.

\section{Related Work}
\label{sec:related}

TLO connects three lines of work: final-response safety evaluation, hidden-state safety representations, and inference-time defenses.
These respectively measure jailbreak outcomes, expose internal safety signals, and intervene during decoding, but none provides a calibrated temporal diagnostic from standard logits.
TLO fills this gap as a passive logit-level observation method that tracks when, and whether, safety evidence becomes detectable during generation.

\paragraph{Safety evaluation.}
Safety training through human preference learning has produced models that resist harmful requests, motivating evaluation frameworks that measure this resistance at scale~\citep{ziegler2019fine, stiennon2020learning}.
Benchmarks such as HarmBench~\citep{mazeika2024harmbench} and AdvBench~\citep{zou2023universal}, together with red-teaming efforts~\citep{ganguli2022red, perez2022red}, standardize jailbreak evaluation largely through ASR.
ASR is practical and widely adopted, but it reduces the decoding process to a final-response label.
TLO complements these outcome-level metrics by tracking the compliance--refusal balance during decoding and calibrating it into RP-plane coordinates, exposing the cross-condition geometry that ASR cannot represent.

\paragraph{Safety representations.}
Representation-level work shows that safety behavior leaves measurable traces in hidden states.
Refusal-direction analysis identifies linear directions associated with refusal, representation engineering uses such directions for behavioral control, and response-safety representations rank outputs by safety quality~\citep{arditi2024refusal, zou2023representation, du2025safetyrepresentation}.
\citet{yin2025refusal} shows that hidden-state refusal scores can change at token-level time scales, but this evidence requires activation access.
\citet{li2025logit} use a compliance--refusal logit gap for steering; we adopt the same observable for passive temporal diagnosis, and contribute the RP calibration that makes the resulting trajectory comparable across models, attacks, and prompt formats.
Activation-based detectors are stronger-access baselines when hidden-state access is available; TLO targets the lower-access full-logit setting that frontier-style auditing increasingly demands.

\paragraph{Inference-time detection and defense.}
Inference-time detection and defense methods use generation-side signals to identify or reduce jailbreak risk without changing model weights.
Early-logit detectors show that first-pass or first-token output distributions can efficiently flag jailbreak inputs~\citep{candogan2025single, chen2025llm}; these methods produce per-prompt verdicts and target input classification.
SafeDecoding~\citep{xu2024safedecoding} re-weights token probabilities toward safe continuations, while CARE~\citep{hu2025care} uses a detect--rollback loop to halt harmful generations.
TLO targets a different task.
Rather than a per-prompt verdict or a decoding policy, it is a calibrated cross-condition diagnostic:
it tracks how compliance--refusal evidence evolves across the full decoding window, and exposes model--attack interactions that per-prompt detectors and uniform decoding policies do not see.
\section{Discussion and Limitations}
\label{sec:Limitations}

TLO offers a calibrated logit-only diagnostic for safety failures.
We state several limits explicitly to scope how it should be interpreted.

\paragraph{Access and coverage.}
TLO requires full logits or targeted logit-bias probing---standard top-$k$ outputs at $k=20$ are insufficient (Appendix~\ref{appendix:topk_truncation})---and uses a fixed English compliance--refusal lexicon supported by probe-derived rank checks, perturbations, and data-driven induction (Appendix~\ref{app:probe_lexicon}, \ref{app:lexicon_ablation}, \ref{app:lexicon_induction}).
It is therefore most directly applicable to open-weight English auditing; per-language or per-model lexicons can be plugged into the same pipeline, while API-restricted deployment via lightweight distilled monitors is left to future work, and adaptive manipulation of the logit channel itself is outside our threat model.

\paragraph{Format effects and observational scope.}
Temporal observability responds to attack formatting as well as harmful intent: format-free benign queries are stable, but attack-formatted benign references can trigger the same temporal cues used to detect attacks, motivating $S_0$-gated variants for deployment-aware monitors (Appendix~\ref{app:s0_gating}).
More broadly, TLO is observational---it makes safety-failure dynamics visible from logits but does not identify underlying circuits or causal mechanisms; this limit is intentional, since observation is what the field currently lacks for ASR-collapsed evaluation.
Future work should connect TLO observations to specific intervention mechanisms, expand to multilingual lexicons and frontier model scales, and develop standardized audit protocols that preserve the observation--causation distinction.

\section{Conclusion}

ASR records whether an attack succeeded but not how.
\emph{Temporal Logit Observability} (TLO) makes the missing dimension observable from logits alone: a training-free 2D RP-plane coordinate, designed to be complementary to ASR rather than to replace it.
Across four aligned LLMs and three jailbreak paradigms, the RP plane separates ASR-neighboring conditions that ASR collapses, agrees with hidden-state refusal directions on 10 of 12 conditions with model-specific scope made explicit, and drives a uniform early-stop rule that cuts aggregate ASR by 26.5\% without false alarms on plain benign queries.
This first step, from observation to actionable monitoring without attack-specific tuning, opens the more direct question of whether attack-aware, RP-aware defenses can do better than the uniform halt rule reported here.
Safety evaluation should report \emph{when} and \emph{how} a failure unfolds, not only \emph{whether} it occurred.

\bibliographystyle{plainnat}
\bibliography{Reference}

@article{vaswani2017attention,
  title={Attention is All You Need},
  author={Vaswani, Ashish and Shazeer, Noam and Parmar, Niki and Uszkoreit, Jakob and Jones, Llion and Gomez, Aidan N. and Kaiser, {\L}ukasz and Polosukhin, Illia},
  journal={Advances in Neural Information Processing Systems},
  volume={30},
  year={2017}
}

@article{brown2020language,
  title={Language Models are Few-Shot Learners},
  author={Brown, Tom B. and Mann, Benjamin and Ryder, Nick and Subbiah, Melanie and Kaplan, Jared and Dhariwal, Prafulla and Neelakantan, Arvind and Shyam, Pranav and Sastry, Girish and Askell, Amanda and others},
  journal={Advances in Neural Information Processing Systems},
  volume={33},
  pages={1877--1901},
  year={2020}
}

@misc{ganguli2022red,
      title={Red Teaming Language Models to Reduce Harms: Methods, Scaling Behaviors, and Lessons Learned}, 
      author={Deep Ganguli and Liane Lovitt and Jackson Kernion and Amanda Askell and Yuntao Bai and Saurav Kadavath and Ben Mann and Ethan Perez and Nicholas Schiefer and Kamal Ndousse and Andy Jones and Sam Bowman and Anna Chen and Tom Conerly and Nova DasSarma and Dawn Drain and Nelson Elhage and Sheer El-Showk and Stanislav Fort and Zac Hatfield-Dodds and Tom Henighan and Danny Hernandez and Tristan Hume and Josh Jacobson and Scott Johnston and Shauna Kravec and Catherine Olsson and Sam Ringer and Eli Tran-Johnson and Dario Amodei and Tom Brown and Nicholas Joseph and Sam McCandlish and Chris Olah and Jared Kaplan and Jack Clark},
      year={2022},
      eprint={2209.07858},
      archivePrefix={arXiv},
      primaryClass={cs.CL},
      url={https://arxiv.org/abs/2209.07858}, 
}

@article{liu2023jailbreaking,
  title={Jailbreaking {ChatGPT} via Prompt Engineering: An Empirical Study},
  author={Liu, Yi and Deng, Gelei and Xu, Zhengzi and Li, Yuekang and Zheng, Yaowen and Zhang, Ying and Zhao, Lida and Zhang, Tianwei and Wang, Kailong and Liu, Yang},
  journal={arXiv preprint arXiv:2305.13860},
  year={2023}
}

@article{li2023deepinception,
  title={Deepinception: Hypnotize large language model to be jailbreaker},
  author={Li, Xuan and Zhou, Zhanke and Zhu, Jianing and Yao, Jiangchao and Liu, Tongliang and Han, Bo},
  journal={arXiv preprint arXiv:2311.03191},
  year={2023}
}

@article{zou2023universal,
  title={Universal and transferable adversarial attacks on aligned language models},
  author={Zou, Andy and Wang, Zifan and Carlini, Nicholas and Nasr, Milad and Kolter, J Zico and Fredrikson, Matt},
  journal={arXiv preprint arXiv:2307.15043},
  year={2023}
}

@article{inan2023llama,
  title={{Llama Guard}: {LLM}-based Input-Output Safeguard for Human-{AI} Conversations},
  author={Inan, Hakan and Upasani, Kartikeya and Chi, Jianfeng and Rungta, Rashi and Iyer, Krithika and Mao, Yuning and Tontchev, Michael and Hu, Qing and Fuller, Brian and Testuggine, Davide and others},
  journal={arXiv preprint arXiv:2312.06674},
  year={2023}
}

@inproceedings{xu2024safedecoding,
  title={{SafeDecoding}: Defending against Jailbreak Attacks via Safety-Aware Decoding},
  author={Xu, Zhangchen and Jiang, Fengqing and Niu, Luyao and Jia, Jinyuan and Lin, Bill Yuchen and Poovendran, Radha},
  booktitle={Proceedings of the 62nd Annual Meeting of the Association for Computational Linguistics (Volume 1: Long Papers)},
  pages={5587--5605},
  year={2024}
}

@article{hu2025care,
  title={{CARE}: Decoding Time Safety Alignment via Rollback and Introspection Intervention},
  author={Hu, Xiaomeng and Huang, Fei and Yuan, Chenhan and Lin, Junyang and Ho, Tsung-Yi},
  journal={arXiv preprint arXiv:2509.06982},
  year={2025}
}

@article{arditi2024refusal,
  title={Refusal in language models is mediated by a single direction},
  author={Arditi, Andy and Obeso, Oscar and Syed, Aaquib and Paleka, Daniel and Panickssery, Nina and Gurnee, Wes and Nanda, Neel},
  journal={Advances in Neural Information Processing Systems},
  volume={37},
  pages={136037--136083},
  year={2024}
}

@article{zou2023representation,
  title={Representation Engineering: A Top-Down Approach to {AI} Transparency},
  author={Zou, Andy and Phan, Long and Chen, Sarah and Campbell, James and Guo, Phillip and Ren, Richard and Pan, Alexander and Yin, Xuwang and Mazeika, Mantas and Dombrowski, Ann-Kathrin and others},
  journal={arXiv preprint arXiv:2310.01405},
  year={2023}
}

@article{du2025safetyrepresentation,
  title={Advancing {LLM} Safe Alignment with Safety Representation Ranking},
  author={Du, Tianqi and Wei, Zeming and Chen, Quan and Zhang, Chenheng and Wang, Yisen},
  journal={arXiv preprint arXiv:2505.15710},
  year={2025}
}

@article{li2025logit,
  title={Logit-Gap Steering: Efficient Short-Suffix Jailbreaks for Aligned Large Language Models},
  author={Li, Tung-Ling and Liu, Hongliang},
  journal={arXiv preprint arXiv:2506.24056},
  year={2025}
}

@article{yin2025refusal,
  title={Refusal Falls off a Cliff: How Safety Alignment Fails in Reasoning?},
  author={Yin, Qingyu and Leong, Chak Tou and Yang, Linyi and Huang, Wenxuan and Li, Wenjie and Wang, Xiting and Yoon, Jaehong and Gu, Jinjin and others},
  journal={arXiv preprint arXiv:2510.06036},
  year={2025}
}

@article{mazeika2024harmbench,
  title={{HarmBench}: A Standardized Evaluation Framework for Automated Red Teaming and Robust Refusal},
  author={Mazeika, Mantas and Phan, Long and Yin, Xuwang and Zou, Andy and Wang, Zifan and Mu, Norman and Sakhaee, Elham and Li, Nathaniel and Basart, Steven and Li, Bo and others},
  journal={arXiv preprint arXiv:2402.04249},
  year={2024}
}

@article{chao2024jailbreakbench,
  title={{JailbreakBench}: An Open Robustness Benchmark for Jailbreaking Large Language Models},
  author={Chao, Patrick and Debenedetti, Edoardo and Robey, Alexander and Andriushchenko, Maksym and Croce, Francesco and Sehwag, Vikash and Dobriban, Edgar and Flammarion, Nicolas and Pappas, George J and Tramer, Florian and others},
  journal={Advances in Neural Information Processing Systems},
  volume={37},
  pages={55005--55029},
  year={2024}
}

@article{grattafiori2024llama3,
  title={The {L}lama 3 Herd of Models},
  author={Grattafiori, Aaron and Dubey, Abhimanyu and Jauhri, Abhinav and Pandey, Abhinav and Kadian, Abhishek and Al-Dahle, Ahmad and Letman, Aiesha and Mathur, Akhil and Schelten, Alan and others},
  journal={arXiv preprint arXiv:2407.21783},
  year={2024}
}

@article{jiang2023mistral,
  title={Mistral {7B}},
  author={Jiang, Albert Q. and Sablayrolles, Alexandre and Mensch, Arthur and Bamford, Chris and Chaplot, Devendra Singh and de las Casas, Diego and Bressand, Florian and Lengyel, Gianna and Lample, Guillaume and Saulnier, Lucile and others},
  journal={arXiv preprint arXiv:2310.06825},
  year={2023}
}

@article{yang2024qwen25,
  title={Qwen2.5 Technical Report},
  author={Yang, An and Yang, Baosong and Zhang, Beichen and Hui, Binyuan and Zheng, Bo and Yu, Bowen and Li, Chengyuan and Liu, Dayiheng and Huang, Fei and others},
  journal={arXiv preprint arXiv:2412.15115},
  year={2024}
}

@inproceedings{christiano2017deep,
  title={Deep Reinforcement Learning from Human Preferences},
  author={Christiano, Paul F. and Leike, Jan and Brown, Tom B. and Martic, Miljan and Legg, Shane and Amodei, Dario},
  booktitle={Advances in Neural Information Processing Systems},
  volume={30},
  year={2017}
}

@article{ziegler2019fine,
  title={Fine-Tuning Language Models from Human Preferences},
  author={Ziegler, Daniel M. and Stiennon, Nisan and Wu, Jeffrey and Brown, Tom B. and Radford, Alec and Amodei, Dario and Christiano, Paul and Irving, Geoffrey},
  journal={arXiv preprint arXiv:1909.08593},
  year={2019}
}

@inproceedings{stiennon2020learning,
  title={Learning to Summarize with Human Feedback},
  author={Stiennon, Nisan and Ouyang, Long and Wu, Jeffrey and Ziegler, Daniel M. and Lowe, Ryan and Voss, Chelsea and Radford, Alec and Amodei, Dario and Christiano, Paul F.},
  booktitle={Advances in Neural Information Processing Systems},
  volume={33},
  pages={3008--3021},
  year={2020}
}

@article{bai2022helpful,
  title={Training a Helpful and Harmless Assistant with Reinforcement Learning from Human Feedback},
  author={Bai, Yuntao and Jones, Andy and Ndousse, Kamal and Askell, Amanda and Chen, Anna and DasSarma, Nova and Drain, Dawn and Fort, Stanislav and Ganguli, Deep and Henighan, Tom and others},
  journal={arXiv preprint arXiv:2204.05862},
  year={2022}
}

@article{bai2022constitutional,
  title={Constitutional {AI}: Harmlessness from {AI} Feedback},
  author={Bai, Yuntao and Kadavath, Saurav and Kundu, Sandipan and Askell, Amanda and Kernion, Jackson and Jones, Andy and Chen, Anna and Goldie, Anna and Mirhoseini, Azalia and McKinnon, Cameron and others},
  journal={arXiv preprint arXiv:2212.08073},
  year={2022}
}

@inproceedings{perez2022red,
  title={Red Teaming Language Models with Language Models},
  author={Perez, Ethan and Huang, Saffron and Song, H. Francis and Cai, Trevor and Ring, Roman and Aslanides, John and Glaese, Amelia and McAleese, Nat and Irving, Geoffrey},
  booktitle={Proceedings of the 2022 Conference on Empirical Methods in Natural Language Processing},
  pages={3419--3448},
  year={2022}
}

@article{ouyang2022training,
  title={Training language models to follow instructions with human feedback},
  author={Ouyang, Long and Wu, Jeffrey and Jiang, Xu and Almeida, Diogo and Wainwright, Carroll and Mishkin, Pamela and Zhang, Chong and Agarwal, Sandhini and Slama, Katarina and Ray, Alex and others},
  journal={Advances in Neural Information Processing Systems},
  volume={35},
  pages={27730--27744},
  year={2022}
}

@article{fawcett2006introduction,
  title={An introduction to {ROC} analysis},
  author={Fawcett, Tom},
  journal={Pattern Recognition Letters},
  volume={27},
  number={8},
  pages={861--874},
  year={2006},
  publisher={Elsevier}
}

@article{gemma2team2024,
  title={Gemma 2: Improving Open Language Models at a Practical Size},
  author={{Gemma Team} and Riviere, Morgane and Pathak, Shreya and Sessa, Pier Giuseppe and Hardin, Cassidy and Bhupatiraju, Surya and Hussenot, L{\'e}onard and Mesnard, Thomas and Shahriari, Bobak and others},
  journal={arXiv preprint arXiv:2408.00118},
  year={2024}
}

@article{chen2025llm,
  title={{LLM} jailbreak detection for (almost) free!},
  author={Chen, Guorui and Xia, Yifan and Jia, Xiaojun and Li, Zhijiang and Torr, Philip and Gu, Jindong},
  journal={arXiv preprint arXiv},
  volume={2509},
  year={2025}
}

@article{candogan2025single,
  title={Single-pass detection of jailbreaking input in large language models},
  author={Candogan, Leyla Naz and Wu, Yongtao and Rocamora, Elias Abad and Chrysos, Grigorios G. and Cevher, Volkan},
  journal={arXiv preprint arXiv:2502.15435},
  year={2025}
}
\newpage
\appendix

\section{Evaluation Grid and Signal Anchors}
\label{app:evaluation_grid}

The evaluation grid supports the main observability claim with 12 model--attack conditions.
We evaluate Llama-3.1-8B-Instruct~\citep{grattafiori2024llama3}, Mistral-7B-Instruct-v0.3~\citep{jiang2023mistral}, Qwen2.5-7B-Instruct~\citep{yang2024qwen25}, and Gemma-2-9B-Instruct~\citep{gemma2team2024} against multi-turn context manipulation (MCM), GCG~\citep{zou2023universal}, and DeepInception (DI)~\citep{li2023deepinception}.
Each condition contains 60 harmful JailbreakBench prompts~\citep{chao2024jailbreakbench} and 60 attack-formatted benign references.
Llama-Guard-3-8B~\citep{inan2023llama} provides the primary success labels; Appendix~\ref{app:judge} records the auxiliary-judge refreshes and their condition-level disagreements.

\begin{table}[h]
    \centering
    \caption{Evaluation grid and calibrated signal anchors. Harmful columns report ASR and condition-mean signal values; benign columns report attack-formatted reference anchors for RP calibration.}
    \label{tab:dataset}
    \small
    \begin{tabular}{llrrrrr}
        \toprule
        & & \multicolumn{3}{c}{Harmful} & \multicolumn{2}{c}{Benign} \\
        \cmidrule(lr){3-5} \cmidrule(lr){6-7}
        Model & Attack & ASR & $S_0$ & $\bar{S}$ & $S_0$ & $\bar{S}$ \\
        \midrule
        \multirow{3}{*}{Llama}   & MCM           &  5.0\% &  1.45 & $-1.42$ &  6.38 & $+1.25$ \\
                                  & GCG           &  8.3\% &  0.62 & $-0.99$ &  4.29 & $-0.53$ \\
                                  & DI & 55.0\% &  4.43 & $-0.11$ &  6.50 & $+0.72$ \\
        \cmidrule(lr){1-7}
        \multirow{3}{*}{Mistral} & MCM           &  5.0\% &  4.51 & $-0.31$ &  5.25 & $+0.67$ \\
                                  & GCG           & 56.7\% &  3.14 & $+0.16$ &  4.76 & $-0.52$ \\
                                  & DI & 80.0\% &  3.04 & $+0.43$ &  3.10 & $+0.69$ \\
        \cmidrule(lr){1-7}
        \multirow{3}{*}{Qwen}    & MCM           & 40.0\% & 11.97 & $-0.12$ & 12.89 & $+1.43$ \\
                                  & GCG           & 76.7\% &  7.13 & $+0.71$ & 12.80 & $-0.60$ \\
                                  & DI & 58.3\% &  7.86 & $+1.18$ & 12.52 & $+1.63$ \\
        \cmidrule(lr){1-7}
        \multirow{3}{*}{Gemma}   & MCM           & 18.3\% &  4.67 & $+0.47$ &  9.74 & $+2.32$ \\
                                  & GCG           & 33.3\% &  4.46 & $+0.68$ &  8.63 & $+1.07$ \\
                                  & DI & 38.3\% &  4.63 & $+0.89$ &  6.69 & $+1.49$ \\
        \bottomrule
    \end{tabular}
\end{table}

\subsection{Authoritative 12-Condition Result Ledger}
\label{app:authoritative_12}

All main-text aggregate results use the latest 12-condition Llama-Guard grid ($n=720$ harmful samples).
This appendix reports the same 12-condition values whenever an aggregate or per-condition number is used as evidence for the main claims.

\begin{table}[h]
    \centering
    \caption{Primary 12-condition success-prediction summary ($n=720$ harmful samples; Llama-Guard success labels).}
    \label{tab:primary_auc_12}
    \small
    \begin{tabular}{@{}lcc@{}}
        \toprule
        Metric & ROC-AUC & 95\% CI \\
        \midrule
        $t_{\text{cross}}$ & 0.784 & [0.750, 0.816] \\
        $\bar{S}_{1{:}5}$ & 0.771 & [0.737, 0.804] \\
        $\bar{S}$ & 0.758 & [0.724, 0.793] \\
        Keyword refusal baseline & 0.748 & [0.715, 0.780] \\
        $\bar{S}_{1{:}3}$ & 0.723 & [0.687, 0.759] \\
        $S_0$ & 0.680 & [0.640, 0.718] \\
        \bottomrule
    \end{tabular}
\end{table}

\begin{table}[h]
    \centering
    \caption{Per-condition AUC heterogeneity in the same 12-condition grid. Values are oriented so larger AUC indicates stronger success/failure separation for that condition.}
    \label{tab:per_condition_auc_12}
    \scriptsize
    \resizebox{\textwidth}{!}{%
    \begin{tabular}{@{}llrrcccccc@{}}
        \toprule
        Model & Attack & $n_{\text{succ}}$ & ASR & $S_0$ & $\bar{S}_{1{:}3}$ & $\bar{S}_{1{:}5}$ & $\bar{S}$ & $t_{\text{cross}}$ & Keyword \\
        \midrule
        Gemma & DI  & 23 & 38.3\% & 0.792 & 0.760 & 0.719 & 0.806 & 0.792 & 0.832 \\
        Gemma & GCG & 20 & 33.3\% & 0.723 & 0.672 & 0.693 & 0.725 & 0.629 & 0.725 \\
        Gemma & MCM & 11 & 18.3\% & 0.700 & 0.688 & 0.696 & 0.857 & 0.528 & 0.510 \\
        Llama & DI  & 33 & 55.0\% & 0.699 & 0.717 & 0.750 & 0.934 & 0.799 & 0.825 \\
        Llama & GCG &  5 &  8.3\% & 0.935 & 0.971 & 0.964 & 0.920 & 0.987 & 0.855 \\
        Llama & MCM &  3 &  5.0\% & 0.924 & 0.918 & 0.918 & 0.977 & 0.614 & 0.649 \\
        Mistral & DI  & 48 & 80.0\% & 0.542 & 0.535 & 0.634 & 0.521 & 0.667 & 0.625 \\
        Mistral & GCG & 34 & 56.7\% & 0.745 & 0.791 & 0.827 & 0.631 & 0.705 & 0.693 \\
        Mistral & MCM &  3 &  5.0\% & 0.737 & 0.725 & 0.708 & 0.942 & 0.570 & 0.947 \\
        Qwen & DI  & 35 & 58.3\% & 0.619 & 0.627 & 0.627 & 0.574 & 0.653 & 0.723 \\
        Qwen & GCG & 46 & 76.7\% & 0.511 & 0.517 & 0.550 & 0.516 & 0.519 & 0.540 \\
        Qwen & MCM & 24 & 40.0\% & 0.536 & 0.532 & 0.520 & 0.595 & 0.685 & 0.500 \\
        \midrule
        \multicolumn{4}{r}{Macro average} & 0.705 & 0.705 & 0.717 & 0.750 & 0.679 & 0.702 \\
        \bottomrule
    \end{tabular}%
    }
\end{table}

The pooled AUC in Table~\ref{tab:primary_auc_12} is a summary over heterogeneous model--attack conditions, not a claim that every condition is equally separable.
Table~\ref{tab:per_condition_auc_12} makes this heterogeneity explicit: Qwen+GCG is near chance for all reported logit summaries, while several Llama conditions are near ceiling.

\subsection{ASR-Neighboring Motivation}
\label{app:asr_collapse}

\begin{figure}[h]
    \centering
    \includegraphics[width=0.95\textwidth]{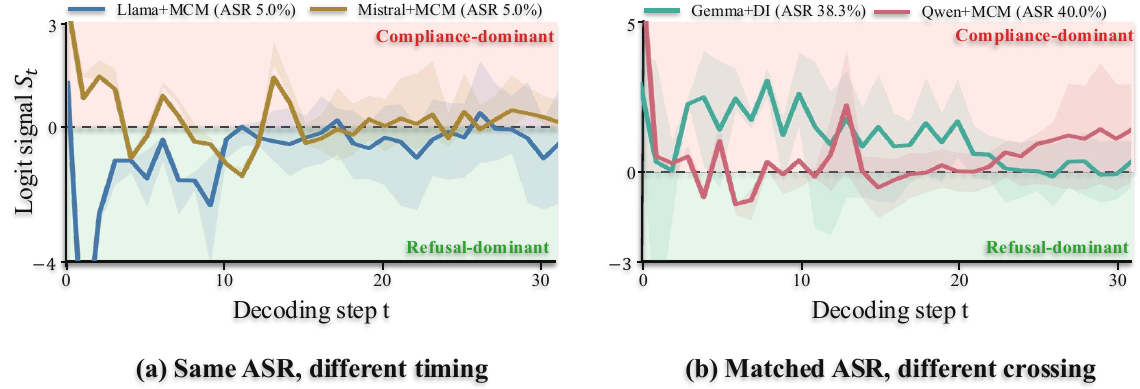}
    \caption{\textbf{ASR-neighboring conditions can have different raw trajectories.}
    Representative condition pairs with identical or near-identical ASR show different LMS crossing behavior.
    These raw trajectories are qualitative motivation only; the main text uses calibrated RP-plane distances for the primary comparison.}
    \label{fig:asr_collapse}
\end{figure}

ASR-neighboring model--attack pairs can be temporally non-equivalent.
Figure~\ref{fig:asr_collapse} retains raw trajectory examples as qualitative motivation.
The main text reports the calibrated RP-plane version of this comparison in Figure~\ref{fig:rp_plane_asr_spectrum} and Appendix~\ref{app:rp_sensitivity_checks}.

\section{Lexicon and Method Details}
\label{app:method_details}

\subsection{Lexicon Specification}
\label{app:lexicon}

The Logit-Margin Score uses a fixed English compliance--refusal lexicon to keep the diagnostic interpretable and training-free.
The lexicon is intentionally small; Appendix~\ref{app:lexicon_ablation} and Appendix~\ref{app:probe_lexicon} test whether this choice drives the conclusions.

\paragraph{Refusal Lexicon ($\mathcal{L}_{\text{ref}}$, 26 entries).}
\begin{quote}
\small
\texttt{%
"I can't", "I cannot", "I won't", "I will not", "I'm unable", "I am unable",
"I can't help", "I can't assist", "I cannot assist",
"I must refuse", "I have to refuse", "I decline", "I must decline",
"Sorry", "I'm sorry", "Apologies", "I apologize",
"policy", "policies", "guidelines", "rules", "terms",
"inappropriate", "illegal", "harmful", "dangerous"
}
\end{quote}

\paragraph{Compliance Lexicon ($\mathcal{L}_{\text{cmp}}$, 21 entries).}
\begin{quote}
\small
\texttt{%
"Sure", "Certainly", "Of course", "Here's", "Below is",
"Step 1", "Step 2", "First", "Second", "Third", "Next",
"Instructions", "Guide", "Tutorial", "Procedure",
"You can", "To do this", "Do the following",
"Detailed", "Comprehensive", "Complete"
}
\end{quote}

\paragraph{Implementation details.}
Surface phrases are expanded to tokenizer-level IDs, including bare and whitespace-prefixed forms, and duplicate IDs are removed.
For multi-token surface phrases such as \texttt{Step 1}, we take the union of all constituent tokenizer IDs produced by these encodings rather than matching an ordered phrase at runtime.
This order-free expansion can include common constituent tokens, such as digits or frequent subword pieces, that also appear in benign contexts.
We treat this as a deliberate simplification rather than a phrase-level matcher: the analyses aggregate over compliance--refusal margins, calibrate those margins against matched harmful and attack-formatted benign references, and remain stable under lexicon perturbations and lexicon-internal top-$k$ aggregation choices (Appendix~\ref{app:lexicon_ablation}; Appendix~\ref{app:topk_aggregation}).
These checks suggest that union-token artifacts are unlikely to explain the observed temporal separation, while phrase-level or order-aware matching remains a useful refinement for future implementations.
At each decoding step $t$, logits are aggregated over each tokenized lexicon using top-$k$ selection within the lexicon with $k=10$:
\begin{align}
\mu_{\text{cmp}}(t) &= \frac{1}{k} \sum_{v \in \text{top-}k(\mathcal{L}_{\text{cmp}})} \ell_t(v), \\
\mu_{\text{ref}}(t) &= \frac{1}{k} \sum_{v \in \text{top-}k(\mathcal{L}_{\text{ref}})} \ell_t(v), \\
S_t &= \mu_{\text{cmp}}(t) - \mu_{\text{ref}}(t).
\end{align}
We use $k=10$ to smooth single-token volatility without averaging over the full tokenized lexicon; Appendix~\ref{app:topk_aggregation} reports sensitivity to this choice.

\subsection{Multi-Turn Context Manipulation Protocol}
\label{app:mcm}

MCM represents multi-turn context-level pressure rather than token-level suffix optimization.
The attack submits four turns: three context-normalizing turns followed by the target harmful request.
Unlike single-prompt simulations of dialogue history, this protocol lets the model's own previous responses become part of the next-turn context.

\subsection{Benign Sample Exclusion Rationale}
\label{app:benign_exclusion}

Sign reversal is computed only for harmful samples because benign sign patterns mainly reflect prompt format rather than safety activation.
For example, MCM benign samples consistently show $\bar{S} > 0$, while GCG benign samples often show negative $\bar{S}$ because their single-turn format differs structurally from MCM.
We therefore use attack-formatted benign references for RP calibration and reserve sign-reversal interpretation for harmful prompts.

\section{RP-Plane Calibration and Coarse Annotations}
\label{app:calibration_labels}

RP calibration is reference-relative rather than a claim that raw LMS values are directly comparable.
Values outside $[0,1]$ indicate that successful attacks fall outside the harmful--benign reference interval for that condition.
Negative RP values are reported as observed inversions rather than forced into a separate causal interpretation.

\subsection{Reference-Anchor Sensitivity}
\label{app:rp_anchor_sensitivity}

The primary RP anchor uses all harmful samples in a condition as $M_{\text{harmful}}$.
This choice avoids requiring failed jailbreak examples, but it also creates a known sensitivity: in high-ASR conditions, successful samples contribute heavily to the harmful anchor and can shrink the apparent displacement between $M_{\text{success}}$ and $M_{\text{harmful}}$.
For this reason, coarse RP-plane annotations should be read as observations under a fixed reference protocol, not as explanations of underlying causes.
The issue is most relevant for high-ASR conditions such as Mistral+DI (80.0\%) and Qwen+GCG (76.7\%), and least relevant for low-ASR conditions where failures dominate the harmful anchor.
\paragraph{High-ASR anchor compression.}
For a fixed model--attack condition whose harmful anchor is computed from all harmful samples, the all-harmful reference can be written as
\[
M_{\mathrm{harmful}}
=
(1-\mathrm{ASR})M_{\mathrm{failed}}
+
\mathrm{ASR}M_{\mathrm{success}}.
\]
Thus,
\[
M_{\mathrm{success}}-M_{\mathrm{harmful}}
=
(1-\mathrm{ASR})(M_{\mathrm{success}}-M_{\mathrm{failed}}).
\]
As ASR increases, successful samples contribute more heavily to the all-harmful anchor and can compress the apparent displacement from the primary harmful reference.
We therefore interpret NGB as no reliable RP-plane displacement under the fixed reference protocol, not as evidence that no safety-related signal exists anywhere.
The main text therefore uses RP-plane coordinates and coarse annotations to expose model--attack heterogeneity under this fixed reference protocol.
Alternative anchors answer different questions.
This anchor dependence does not imply that one anchor is universally correct; rather, each anchor defines a different comparison question.
A failed-only harmful anchor would isolate successful attacks from observed failures, but it is unstable in high-ASR conditions and unavailable when a condition has too few failed samples.
A format-free benign anchor would estimate deployment-like benign behavior, but would remove the attack wrapper that RP is designed to control.
We therefore use attack-formatted benign references for RP calibration and report format-free benign queries separately as a false-positive characterization.

\subsection{RP-Plane Sensitivity Checks}
\label{app:rp_sensitivity_checks}

RP-plane sensitivity checks test whether calibrated coordinates are driven by denominator instability, ASR-neighbor sampling noise, anchor choice, or raw-logit scale.
All checks use the same 12-condition Llama-Guard primary grid as the main text.
The primary filter reproduces the condition-level ASR, $\mathrm{RP}_A$, $\mathrm{RP}_B$, and $r$ values in Table~\ref{tab:type_ab}.

\paragraph{Denominator diagnostics.}
The RP denominator is stable in nearly all condition axes.
Table~\ref{tab:denominator_diagnostics} reports $\Delta_A=M_{\mathrm{benign}}(S_0)-M_{\mathrm{harmful}}(S_0)$ and $\Delta_B=M_{\mathrm{benign}}(\bar{S})-M_{\mathrm{harmful}}(\bar{S})$.
Only one condition-axis, Mistral+DI on $\Delta_A$, falls below the near-zero threshold $|\Delta|<0.10$.

\begin{table}[h]
    \centering
    \caption{Denominator sign and magnitude audit for RP calibration. Values are benign-minus-harmful anchor gaps; ``Flag'' marks $|\Delta|<0.10$.}
    \label{tab:denominator_diagnostics}
    \small
    \begin{tabular}{lrrc}
        \toprule
        Condition & $\Delta_A$ ($S_0$) & $\Delta_B$ ($\bar{S}$) & Flag \\
        \midrule
        Gemma + DI & 2.06 & 0.60 & -- \\
        Gemma + GCG & 4.16 & 0.39 & -- \\
        Gemma + MCM & 5.06 & 1.84 & -- \\
        Llama + DI & 2.06 & 0.83 & -- \\
        Llama + GCG & 3.67 & 0.46 & -- \\
        Llama + MCM & 4.93 & 2.68 & -- \\
        Mistral + DI & 0.06 & 0.26 & $\Delta_A$ \\
        Mistral + GCG & 1.62 & $-$0.68 & -- \\
        Mistral + MCM & 0.74 & 0.98 & -- \\
        Qwen + DI & 4.65 & 0.45 & -- \\
        Qwen + GCG & 5.67 & $-$1.32 & -- \\
        Qwen + MCM & 0.92 & 1.55 & -- \\
        \bottomrule
    \end{tabular}
\end{table}

\paragraph{ASR-neighboring pair bootstrap.}
Pairwise RP-plane distances use stratified bootstrap resampling within each condition's success, failure, and benign groups, holding the observed ASR and success count fixed.
This provides descriptive CIs for $D_{\mathrm{RP}}$, not a pairwise permutation null distribution.

\begin{table}[h]
    \centering
    \caption{ASR-neighboring condition pairs ($|\Delta\mathrm{ASR}|\leq5$ percentage points) with stratified-bootstrap 95\% CIs for $D_{\mathrm{RP}}$ ($B=10{,}000$).}
    \label{tab:drp_neighbor_ci}
    \scriptsize
    \begin{tabular}{lccc}
        \toprule
        Pair & ASR diff. (pp) & $D_{\mathrm{RP}}$ [95\% CI] & $n_{\mathrm{succ}}$ \\
        \midrule
        Llama+GCG vs. Llama+MCM & 3.3 & 3.156 [2.028, 6.377] & 5/3 \\
        Llama+GCG vs. Mistral+MCM & 3.3 & 3.059 [1.926, 6.256] & 5/3 \\
        Llama+DI vs. Mistral+GCG & 1.7 & 1.014 [0.726, 1.351] & 33/34 \\
        Gemma+DI vs. Qwen+MCM & 1.7 & 0.625 [0.376, 1.131] & 23/24 \\
        Llama+DI vs. Qwen+DI & 3.3 & 0.572 [0.237, 1.026] & 33/35 \\
        Gemma+DI vs. Gemma+GCG & 5.0 & 0.508 [0.296, 1.041] & 23/20 \\
        Mistral+GCG vs. Qwen+DI & 1.7 & 0.508 [0.188, 0.885] & 34/35 \\
        Llama+MCM vs. Mistral+MCM & 0.0 & 0.330 [0.078, 0.851] & 3/3 \\
        Mistral+DI vs. Qwen+GCG & 3.3 & 0.184 [0.033, 1.868] & 48/46 \\
        \midrule
        Median across nine pairs & -- & 0.572 [0.489, 0.924] & -- \\
        \bottomrule
    \end{tabular}
\end{table}

\paragraph{Anchor sensitivity.}
Failed-only harmful anchors preserve raw axis dominance in 11 of 12 conditions, while model-global anchors change raw axis dominance in 5 of 12 conditions.
This supports the condition-relative anchor as the primary coordinate system and treats model-global anchoring as a sensitivity analysis.

\begin{table}[h]
    \centering
    \caption{Anchor sensitivity of RP coordinates. Each cell reports $(\mathrm{RP}_A,\mathrm{RP}_B)$/dominant raw axis before applying NGB filtering.}
    \label{tab:anchor_sensitivity}
    \scriptsize
    \resizebox{\textwidth}{!}{%
    \begin{tabular}{lccc}
        \toprule
        Condition & Primary condition anchor & Failed-only harmful anchor & Model-global anchor \\
        \midrule
        Gemma + DI & (0.72, 0.55)/A & (0.80, 0.66)/A & (0.40, 0.57)/B \\
        Gemma + GCG & (0.27, 0.79)/B & (0.35, 0.85)/B & (0.26, 0.32)/B \\
        Gemma + MCM & (0.09, 0.22)/B & (0.10, 0.26)/B & (0.14, 0.21)/B \\
        Llama + DI & (0.35, 0.84)/B & (0.55, 0.92)/B & (0.84, 1.08)/B \\
        Llama + GCG & (1.01, 3.80)/B & (1.01, 3.08)/B & (0.61, 1.21)/B \\
        Llama + MCM & (0.79, 0.66)/A & (0.80, 0.67)/A & (0.90, 0.89)/A \\
        Mistral + DI & ($-$0.18, 0.00)/A & ($-$3.10, 0.01)/A & ($-$0.66, 1.82)/B \\
        Mistral + GCG & (0.30, $-$0.17)/A & (0.50, $-$0.51)/B & (0.08, 0.98)/B \\
        Mistral + MCM & (0.49, 0.79)/B & (0.50, 0.80)/B & (1.62, 1.96)/B \\
        Qwen + DI & (0.14, 0.31)/B & (0.28, 0.52)/B & ($-$0.13, 3.18)/B \\
        Qwen + GCG & (0.01, 0.00)/A & (0.02, 0.00)/A & ($-$0.49, 0.53)/B \\
        Qwen + MCM & (0.32, 0.07)/A & (0.43, 0.11)/A & (0.87, $-$2.63)/B \\
        \bottomrule
    \end{tabular}%
    }
\end{table}

\paragraph{Rank-based RP.}
A rank-based RP variant replaces raw LMS scale with percentile positions within the harmful-plus-benign reference distribution for each condition.
The rank-based coordinates strongly track the primary coordinates: Pearson $r=0.858$ for $\mathrm{RP}_A$, $r=0.920$ for $\mathrm{RP}_B$, and $r=0.925$ for displacement magnitude.

\begin{table}[h]
    \centering
    \caption{Rank-based RP sensitivity check. Rank RP uses empirical percentile positions rather than raw LMS scale.}
    \label{tab:rank_based_rp}
    \scriptsize
    \begin{tabular}{lccc}
        \toprule
        Condition & Primary $(\mathrm{RP}_A,\mathrm{RP}_B)$ & Rank-based $(\mathrm{RP}_A,\mathrm{RP}_B)$ & $r_{\mathrm{rank}}$ \\
        \midrule
        Gemma + DI & (0.72, 0.55) & (0.24, 0.33) & 0.41 \\
        Gemma + GCG & (0.27, 0.79) & (0.20, 0.72) & 0.75 \\
        Gemma + MCM & (0.09, 0.22) & (0.16, 0.29) & 0.33 \\
        Llama + DI & (0.35, 0.84) & (0.07, 1.23) & 1.24 \\
        Llama + GCG & (1.01, 3.80) & (0.99, 2.01) & 2.24 \\
        Llama + MCM & (0.79, 0.66) & (0.49, 0.45) & 0.67 \\
        Mistral + DI & ($-$0.18, 0.00) & ($-$0.19, 0.00) & 0.19 \\
        Mistral + GCG & (0.30, $-$0.17) & (0.25, $-$0.16) & 0.30 \\
        Mistral + MCM & (0.49, 0.79) & (0.66, 0.75) & 1.00 \\
        Qwen + DI & (0.14, 0.31) & (0.06, 0.09) & 0.11 \\
        Qwen + GCG & (0.01, 0.00) & (0.01, $-$0.01) & 0.01 \\
        Qwen + MCM & (0.32, 0.07) & (0.20, 0.06) & 0.21 \\
        \bottomrule
    \end{tabular}
\end{table}

\subsection{Permutation Test for Null-generation Bias}
\label{app:permutation_test}

The success-label permutation test decides whether a calibrated displacement is reliably nonzero before assigning a coarse RP-plane annotation.
Within each model--attack condition, we permute success/failure labels, recompute $r=\sqrt{\text{RP}_A^2+\text{RP}_B^2}$, and estimate one-sided permutation $p$-values with $B=10{,}000$ permutations.
The primary family contains the 12 Llama-Guard model--attack conditions, and Benjamini--Hochberg FDR correction is applied at $q=0.05$.
Conditions that do not survive correction receive the Null-generation Bias annotation, meaning the calibrated trajectory summaries provide no reliable evidence for either pre-generation or in-generation bias.
The rerun preserves seven reliable RP-displacement conditions, corresponding to the non-NGB cells in Table~\ref{tab:type_ab}.

\begin{table}[h]
    \centering
    \caption{Permutation test rerun for calibrated displacement $r$. The PGB/IGB/NGB annotations use FDR-corrected significance followed by axis dominance.}
    \label{tab:perm_test}
    \scriptsize
    \begin{tabular}{lccccl}
        \toprule
        Model + Attack & $r_{\text{obs}}$ & $p_{\text{perm}}$ & $p_{\mathrm{BH}}$ & FDR result & Annotation status \\
        \midrule
        Gemma + DI & 0.90 & ${<}.001$ & ${<}.001$ & Significant & Axis-dominant \\
        Gemma + GCG & 0.83 & $.002$ & $.004$ & Significant & Axis-dominant \\
        Gemma + MCM & 0.24 & ${<}.001$ & $.002$ & Significant & Axis-dominant \\
        Llama + DI & 0.91 & ${<}.001$ & ${<}.001$ & Significant & Axis-dominant \\
        Llama + GCG & 3.94 & ${<}.001$ & ${<}.001$ & Significant & Axis-dominant \\
        Llama + MCM & 1.03 & $.001$ & $.002$ & Significant & Axis-dominant \\
        Mistral + DI & 0.18 & $.500$ & $.546$ & Not significant & Null-generation Bias \\
        Mistral + GCG & 0.35 & ${<}.001$ & ${<}.001$ & Significant & Axis-dominant \\
        Mistral + MCM & 0.93 & $.054$ & $.081$ & Not significant & Null-generation Bias \\
        Qwen + DI & 0.34 & $.070$ & $.093$ & Not significant & Null-generation Bias \\
        Qwen + GCG & 0.01 & $.988$ & $.988$ & Not significant & Null-generation Bias \\
        Qwen + MCM & 0.32 & $.280$ & $.336$ & Not significant & Null-generation Bias \\
        \bottomrule
    \end{tabular}
\end{table}

Bootstrap intervals provide a descriptive stability check for calibrated displacement magnitudes.
They complement the primary permutation test but do not override its FDR-corrected annotation decision.
Intervals for very small $n_{\text{succ}}$ are descriptive only and are not used for annotation assignment.

\begin{table}[h]
    \centering
    \caption{Bootstrap confidence intervals for selected calibrated displacement magnitudes $r$. PGB/IGB/NGB annotations follow the primary permutation test and axis-dominance rule.}
    \label{tab:r_bootstrap}
    \small
    \begin{tabular}{lcccc}
        \toprule
        Model + Attack & $n_{\text{succ}}$ & $r$ & 95\% CI & Annotation \\
        \midrule
        Gemma + DI & 23 & 0.90 & [0.73, 1.08] & PGB \\
        Gemma + GCG & 20 & 0.83 & [0.34, 1.41] & IGB \\
        Gemma + MCM & 11 & 0.24 & [0.12, 0.33] & IGB \\
        Llama + DI & 33 & 0.91 & [0.55, 1.21] & IGB \\
        Llama + GCG & 5 & 3.94 & [3.05, 5.32] & IGB \\
        Llama + MCM & 3 & 1.03 & [0.95, 1.09] & PGB \\
        Mistral + DI & 48 & 0.18 & [0.07, 1.37] & NGB \\
        Mistral + GCG & 34 & 0.35 & [0.17, 0.57] & PGB$^\ddagger$ \\
        Mistral + MCM & 3 & 0.93 & [0.44, 1.43] & NGB$^\S$ \\
        Qwen + DI & 35 & 0.34 & [0.09, 0.68] & NGB$^\P$ \\
        Qwen + GCG & 46 & 0.01 & [0.02, 0.22] & NGB$^\dagger$ \\
        Qwen + MCM & 24 & 0.32 & [0.05, 0.81] & NGB \\
        \bottomrule
    \end{tabular}
    \\[2pt]
    \footnotesize{$^\dagger$Qwen+GCG has a near-zero point estimate and non-significant permutation test, but its bootstrap interval is unstable; we therefore interpret it as no reliable RP-axis displacement, not as a proven absence of all safety evidence. $^\ddagger$Mistral+GCG has an inverted generation-time RP value. $^\S$Mistral+MCM shows large displacement magnitude but does not pass the primary permutation test. $^\P$Qwen+DI is borderline under the raw permutation test and does not survive the primary significance stage.}
\end{table}

\subsection{Axis-Dominance and Small-Sample Caveats}
\label{app:axis_stability}

Coarse RP-plane annotations are compact summaries of axis structure, not independently established causes.
They are most fragile when the number of successful attacks is small or when the axis-dominance margin between $|\text{RP}_A|$ and $|\text{RP}_B|$ is close.
After the significance stage, axis dominance is assigned by the larger absolute RP coordinate:
\begin{equation}
    \text{PGB}: |\text{RP}_A| > |\text{RP}_B|, \quad
    \text{IGB}: |\text{RP}_A| < |\text{RP}_B|.
    \label{eq:rp_classify}
\end{equation}
Axis-indeterminate exact ties are reported as N/A.
Table~\ref{tab:axis_margin} reports the raw axis margin used after the significance stage.

\begin{table}[h]
    \centering
    \caption{Axis-dominance margins in the 12-condition grid. Positive margin favors PGB; negative margin favors IGB. NGB annotations override axis dominance when the permutation test is not significant.}
    \label{tab:axis_margin}
    \scriptsize
    \begin{tabular}{@{}llrrrcl@{}}
        \toprule
        Model & Attack & $n_{\text{succ}}$ & $\text{RP}_A$ & $\text{RP}_B$ & $|\text{RP}_A|-|\text{RP}_B|$ & Reading \\
        \midrule
        Gemma & DI & 23 & 0.72 & 0.55 &  0.17 & PGB \\
        Gemma & GCG & 20 & 0.27 & 0.79 & -0.52 & IGB \\
        Gemma & MCM & 11 & 0.09 & 0.22 & -0.13 & IGB, small margin \\
        Llama & DI & 33 & 0.35 & 0.84 & -0.49 & IGB \\
        Llama & GCG & 5 & 1.01 & 3.80 & -2.79 & IGB, low $n_{\text{succ}}$ \\
        Llama & MCM & 3 & 0.79 & 0.66 &  0.14 & PGB, low $n_{\text{succ}}$ \\
        Mistral & DI & 48 & -0.18 & 0.00 &  0.18 & NGB by permutation \\
        Mistral & GCG & 34 & 0.30 & -0.17 &  0.13 & PGB, inverted/small margin \\
        Mistral & MCM & 3 & 0.49 & 0.79 & -0.30 & NGB by permutation, low $n_{\text{succ}}$ \\
        Qwen & DI & 35 & 0.14 & 0.31 & -0.17 & NGB by permutation, borderline \\
        Qwen & GCG & 46 & 0.01 & 0.00 &  0.00 & NGB \\
        Qwen & MCM & 24 & 0.32 & 0.07 &  0.25 & NGB by permutation \\
        \bottomrule
    \end{tabular}
\end{table}

The most cautious reading is therefore condition-level and observational: PGB/IGB annotations indicate which calibrated axis dominates under the primary pipeline, while NGB means no reliable RP-axis displacement after the permutation/FDR stage.

\subsection{Condition-Level RP-Plane Annotations and Behavioral Probe Response}
\label{app:condition_labels}

Table~\ref{tab:type_ab} in the main text reports the per-condition RP-plane coordinates, coarse PGB/IGB/NGB annotations, and behavioral-probe response.
The annotation column is an observational summary of calibrated logit evidence, not a causal explanation.
The behavioral-probe columns are included only to show whether the coarse annotations correspond to different behavioral responses.

\subsection{Entropy Check for Null-generation Bias}
\label{app:entropy_crossval}

Entropy provides an independent check for whether a Null-generation Bias annotation is merely a lexicon artifact.
For Qwen+GCG, generation-time entropy shows almost no differentiation between successful attacks and the harmful baseline ($|d|=0.077$), matching the near-zero calibrated LMS displacement.
For contrast, Llama+MCM shows measurable entropy differentiation ($|d|=0.103$--$0.416$), showing that entropy can detect condition-level variation when such variation is present.

\begin{table}[h]
    \centering
    \caption{Entropy cross-validation for selected conditions.}
    \small
    \begin{tabular}{@{}lcccc@{}}
        \toprule
        Condition & ASR & Entropy $|d|$ & Lexicon $|d|$ & Reading \\
        \midrule
        Qwen + GCG & 76.7\% & 0.077 & 0.005--0.011 & No detectable entropy/LMS displacement \\
        Llama + MCM & 5.0\% & 0.103--0.416 & 0.217--0.256 & Weak-to-medium displacement \\
        \bottomrule
    \end{tabular}
\end{table}

\subsection{Intervention-Relevance Implications}
\label{app:defense_implications}

The coarse RP-plane annotations suggest different intervention-facing readings without claiming hidden causes.
IGB conditions are the most natural fit for generation-time monitors because the discriminative signal appears during decoding.
PGB conditions instead suggest that calibrated evidence is already present at the pre-generation position.
NGB conditions should be treated as observability boundaries: the current calibrated trajectory summaries do not provide reliable evidence for either axis.

\section{Temporal Evidence Supporting Success Prediction}
\label{app:temporal_evidence}

\subsection{Token Rank Linkage}
\label{app:token_rank_linkage}

Token-rank analysis verifies that LMS reflects actual next-token competition.
When $S_t<0$, the highest-ranked refusal token typically appears near the top of the vocabulary; when $S_t>0$, refusal tokens drop by orders of magnitude.
Successful jailbreaks concentrate in compliance-dominant regions where refusal tokens remain low-ranked, supporting the main-text interpretation that delayed or absent refusal dominance is visible in the $S_t$ trajectory.

\begin{figure}[h]
    \centering
    \includegraphics[width=\textwidth]{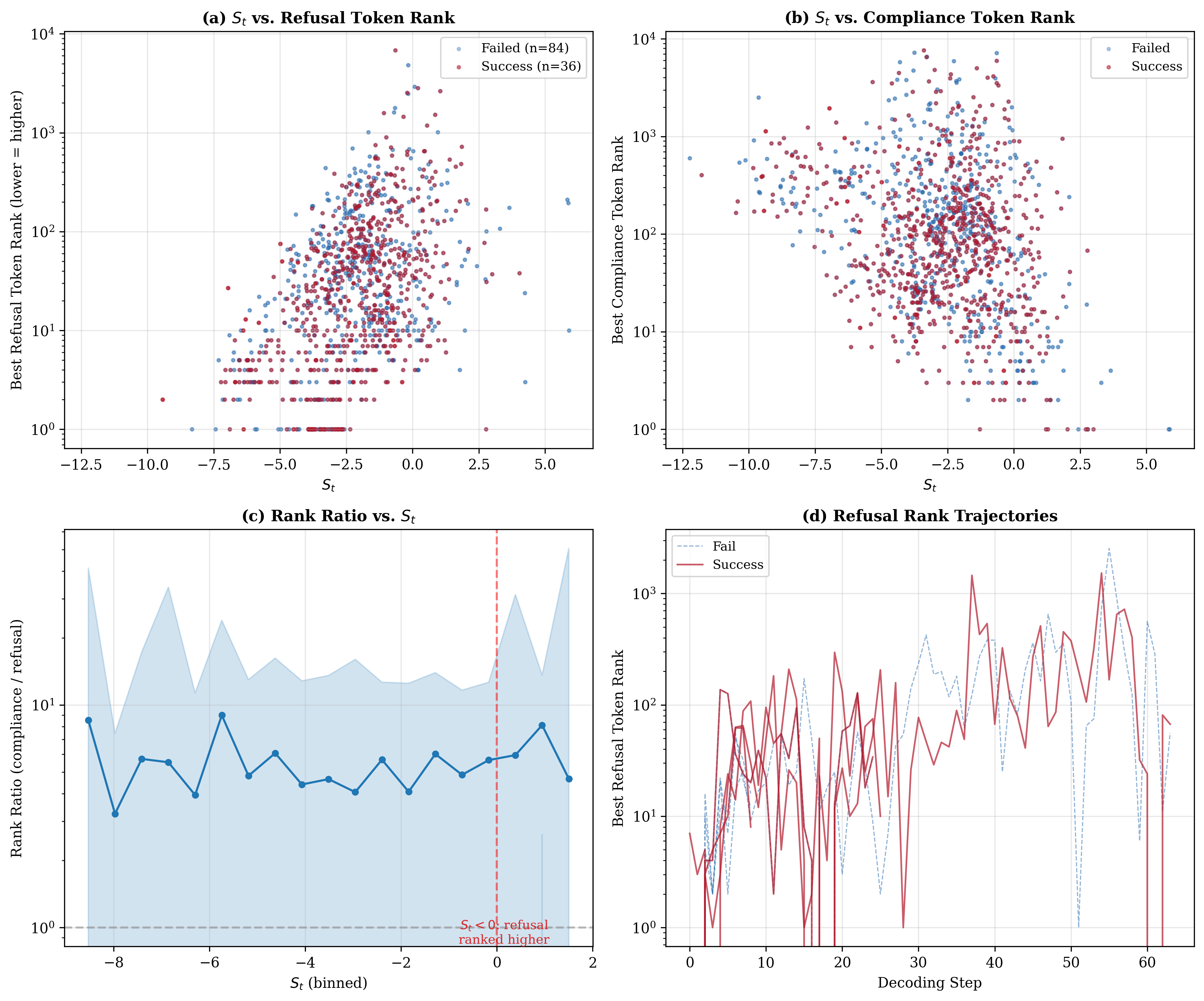}
    \caption{\textbf{Token rank linkage.}
    Negative $S_t$ corresponds to high-ranked refusal tokens, while positive $S_t$ pushes refusal tokens lower in the vocabulary ranking.}
    \label{fig:token_rank}
\end{figure}

\section{Hidden-State References}
\label{app:hidden_crossmodel}

Hidden-state references define the access--observability boundary of the $S_t$ trajectory.
Hidden-state probes use internal activations, so their higher predictive capacity is expected under stronger access assumptions.
The relevant question is where the lower-access $S_t$ trajectory preserves temporally structured evidence and where it weakens.

\subsection{Direction Construction}

The refusal-direction projection uses a per-model, stronger-access hidden-state construct-validity reference rather than an explanatory label.
For each model $m$, we fix the closest layer to 50\% network depth before any alignment evaluation (L50; Llama/Mistral layer 16, Qwen layer 14, Gemma layer 21).
This gives a pre-specified midpoint reference rather than an outcome-tuned hidden-state probe.
At this fixed layer, we compute a difference-of-means direction from paired harmful and benign reference prompts at the final prompt token:
\begin{equation}
    \mathbf{d}_m =
    \frac{\bar{\mathbf{h}}_{\mathrm{harmful},m} - \bar{\mathbf{h}}_{\mathrm{benign},m}}
    {\left\|\bar{\mathbf{h}}_{\mathrm{harmful},m} - \bar{\mathbf{h}}_{\mathrm{benign},m}\right\|_2}.
\end{equation}
The direction uses prompt-reference category rather than jailbreak success labels, and is then held fixed while each decoding-time hidden state is projected as $z_t=\langle \mathbf{h}_t,\mathbf{d}_m\rangle$.
In high-ASR conditions, this reference can also encode harmful-versus-benign context, so we treat it as a construct-validity reference rather than a pure refusal axis.
Because higher $S_t$ means compliance-dominant logits while higher $z_t$ means movement toward the harmful/refusal reference direction, the expected temporal association is negative.

\begin{table}[h]
    \centering
    \caption{\textbf{Condition-level temporal alignment between the $S_t$ trajectory and refusal-direction projections.}
    Attack rows use the same order within each model: MCM, GCG, and DI.}
    \label{tab:hidden_temporal_alignment_full}
    \scriptsize
    \setlength{\tabcolsep}{3.3pt}
    \renewcommand{\arraystretch}{1.05}
    \begin{tabular}{@{}llccccl@{}}
        \toprule
        & & \multicolumn{3}{c}{Temporal association} & \multicolumn{2}{c}{Interpretation} \\
        \cmidrule(lr){3-5} \cmidrule(lr){6-7}
        Model & Attack & $\overline{|\rho|}_{0:4}$ & $\bar{\rho}_{0:4}$ & $\tilde{\rho}_{\mathrm{lag}}$ & Agree. & Reading \\
        \midrule
        \multirow{3}{*}{Llama} & MCM & 0.698 & -0.698 & 0.599 & 0.758 & \textbf{Strong alignment} \\
        & GCG & 0.780 & -0.780 & 0.607 & 0.719 & \textbf{Strong alignment} \\
        & DI  & 0.794 & -0.794 & 0.298 & 0.655 & \textbf{Strong alignment} \\
        \addlinespace[1pt]
        \multirow{3}{*}{Qwen} & MCM & 0.330 &  0.330 & 0.194 & 0.532 & \textit{Boundary} \\
        & GCG & 0.312 & -0.312 & 0.227 & 0.462 & Weak alignment \\
        & DI  & 0.268 &  0.167 & 0.214 & 0.504 & \textit{Boundary} \\
        \addlinespace[1pt]
        \multirow{3}{*}{Mistral} & MCM & 0.177 & -0.174 & 0.196 & 0.531 & Weak alignment \\
        & GCG & 0.460 & -0.460 & 0.272 & 0.572 & Moderate alignment \\
        & DI  & 0.196 & -0.137 & 0.208 & 0.579 & Weak alignment \\
        \addlinespace[1pt]
        \multirow{3}{*}{Gemma} & MCM & 0.217 & -0.209 & 0.184 & 0.551 & Weak alignment \\
        & GCG & 0.384 & -0.384 & 0.241 & 0.615 & Moderate alignment \\
        & DI  & 0.501 & -0.304 & 0.207 & 0.624 & Moderate alignment \\
        \bottomrule
    \end{tabular}
\end{table}

$\overline{|\rho|}_{0:4}$ measures early-step association strength as the mean absolute correlation between $S_t$ and $z_t$ over steps 0--4, $\bar{\rho}_{0:4}$ preserves its direction, $\tilde{\rho}_{\mathrm{lag}}$ measures within-trajectory similarity after lag optimization, and Agree. measures state-sign agreement.

\subsection{Layer-Sweep Sensitivity}
\label{app:hidden_layer_sweep}

The fixed L50 layer is not the only layer at which the temporal alignment appears.
To check whether the hidden-state reference result depends on the midpoint layer choice, we repeat the direction-construction and temporal-alignment analysis at three pre-specified relative depths: L25, L50, and L75.
For each model and layer, the direction is recomputed from the same harmful-versus-benign reference categories at that layer and then held fixed while decoding-time hidden states are projected.
We then summarize the harmful attack conditions (MCM, GCG, DI) using the same early-step association statistic, $\overline{|\rho|}_{0:4}$.

The qualitative pattern is stable across layers.
Llama remains strongly aligned at all three depths, Qwen remains the main boundary case, and Mistral/Gemma retain attack-dependent moderate alignment rather than collapsing at non-L50 layers.
The L50 protocol is therefore a fixed midpoint reference, not a layer chosen to maximize the reported alignment.

\begin{figure}[h]
    \centering
    \includegraphics[width=0.98\textwidth]{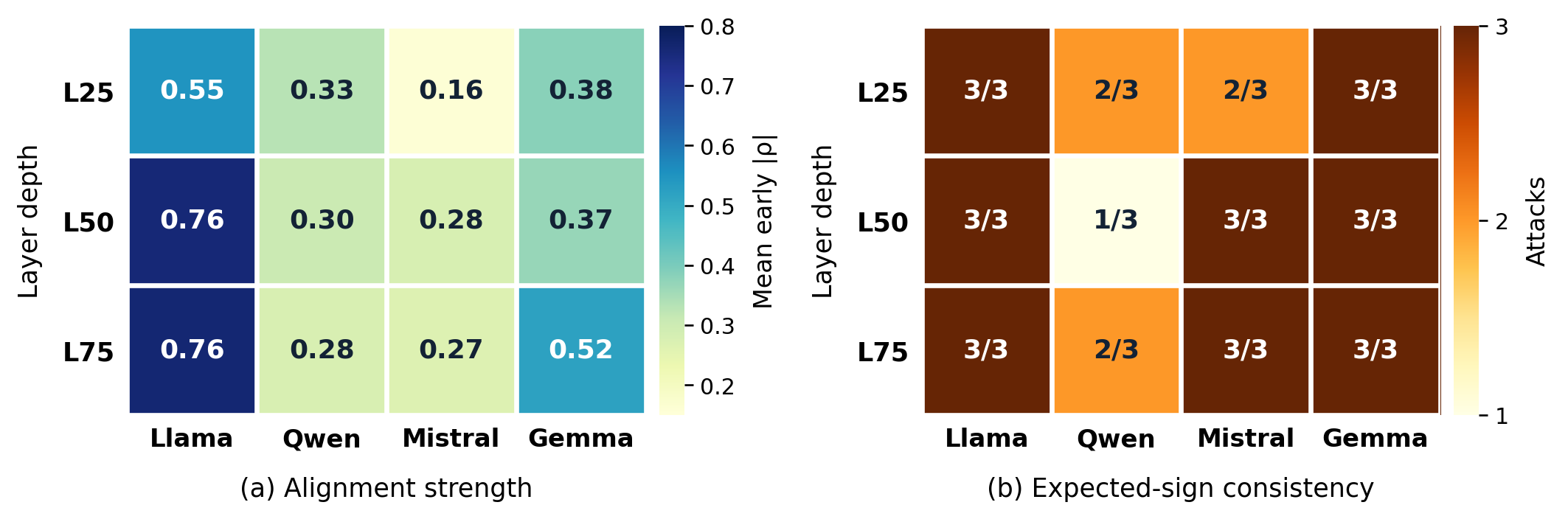}
    \caption{\textbf{Layer-sweep hidden-state alignment.}
    Panel~(a) shows mean early $|\rho|$ between $S_t$ and $z_t$; panel~(b) shows expected-sign consistency across attacks.
    Llama is consistently strongest, while Qwen is the main boundary case.}
    \label{fig:hidden_layer_sweep_main}
\end{figure}

\begin{table}[h]
    \centering
    \caption{\textbf{Hidden-state layer-sweep sensitivity.}
    Entries report the mean $\overline{|\rho|}_{0:4}$ across the three harmful attack conditions at each pre-specified relative depth.
    Parentheses report the number of harmful conditions with the expected negative signed association.}
    \label{tab:hidden_layer_sweep}
    \small
    \begin{tabular}{lccc}
        \toprule
        Model & L25 & L50 & L75 \\
        \midrule
        Llama   & 0.548 (3/3) & 0.757 (3/3) & 0.763 (3/3) \\
        Qwen    & 0.330 (2/3) & 0.303 (1/3) & 0.275 (2/3) \\
        Mistral & 0.159 (2/3) & 0.278 (3/3) & 0.266 (3/3) \\
        Gemma   & 0.382 (3/3) & 0.367 (3/3) & 0.520 (3/3) \\
        \bottomrule
    \end{tabular}
\end{table}

\begin{figure}[h]
    \centering
    \includegraphics[width=0.92\textwidth]{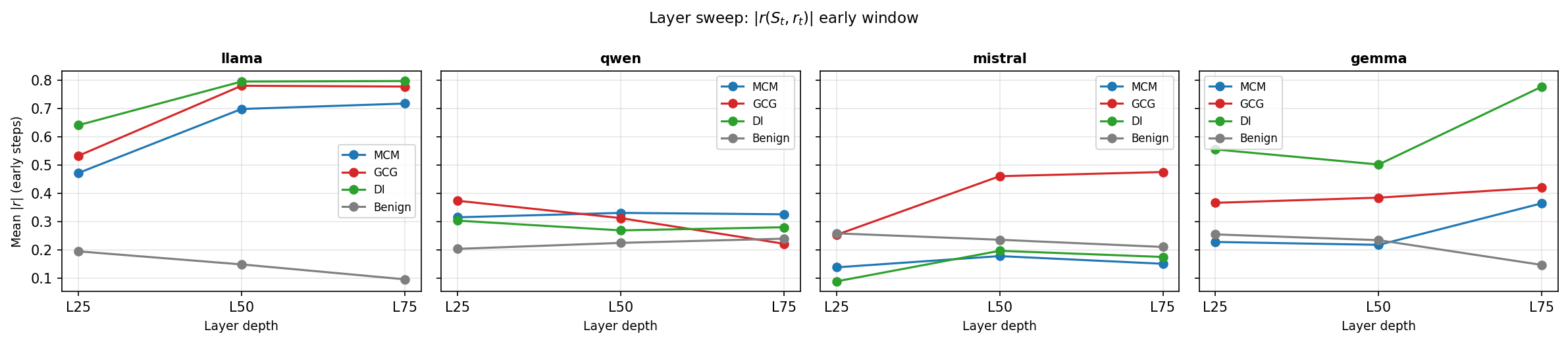}
    \caption{\textbf{Layer-sweep visualization for hidden-state alignment.}
    The plot shows early temporal association between the $S_t$ trajectory and hidden-state reference projections across L25, L50, and L75.
    The same broad interpretation holds away from L50: alignment is strongest and most consistent for Llama, weaker and boundary-like for Qwen, and condition-dependent for Mistral and Gemma.}
    \label{fig:hidden_layer_sweep}
\end{figure}

\begin{table}[h]
\centering
\caption{Success-prediction reference values for $S_t$ summaries and hidden-state probes. Hidden-state probes use stronger access assumptions and serve as construct-validity references; $S_t$ summaries report hidden-state-free logit information.}
\label{tab:hidden_crossmodel_full}
\small
\begin{tabular}{@{}llccccc@{}}
\toprule
Family & Method & Llama & Qwen & Mistral & Gemma & Avg \\
\midrule
\multirow{6}{*}{$S_t$ summaries} & $S_0$ & 0.852 & 0.555 & 0.647 & 0.738 & 0.698 \\
 & $\bar{S}_{1:1}$ & 0.853 & 0.556 & 0.648 & 0.739 & 0.699 \\
 & $\bar{S}_{1:3}$ & 0.874 & 0.570 & 0.667 & 0.667 & 0.694 \\
 & $\bar{S}_{1:5}$ & 0.883 & 0.574 & 0.634 & 0.683 & 0.694 \\
 & $\bar{S}$ & 0.943 & 0.551 & 0.684 & 0.796 & 0.744 \\
 & sign flips & 0.894 & 0.572 & 0.772 & 0.679 & 0.729 \\
\midrule
\multirow{4}{*}{Hidden dir.} & direction, full & 0.874 & 0.728 & 0.905 & 0.797 & 0.826 \\
 & direction, $k=1$ & 0.862 & 0.573 & 0.699 & 0.765 & 0.725 \\
 & direction, $k=3$ & 0.865 & 0.598 & 0.756 & 0.785 & 0.751 \\
 & direction, $k=5$ & 0.861 & 0.588 & 0.770 & 0.793 & 0.753 \\
\midrule
\multirow{4}{*}{Hidden LogReg} & logreg, full & 0.779 & 0.733 & 0.842 & 0.804 & 0.790 \\
 & logreg, $k=1$ & 0.866 & 0.717 & 0.693 & 0.782 & 0.765 \\
 & logreg, $k=3$ & 0.800 & 0.751 & 0.723 & 0.843 & 0.779 \\
 & logreg, $k=5$ & 0.855 & 0.730 & 0.796 & 0.829 & 0.803 \\
\bottomrule
\end{tabular}
\end{table}

The table shows that both $S_t$ summaries and hidden-state probes carry above-chance success-label information across the model grid.
Hidden-state probes generally have higher mean AUC under stronger access assumptions, while $S_t$ summaries remain informative without activation access.
This supports the main claim of the Temporal Logit Observability (TLO) protocol: it preserves an accessible subset of safety-related temporal evidence rather than replacing hidden-state probes.

\section{Robustness and Ablations}
\label{app:robustness_appendix}

\subsection{Stochastic Decoding}
\label{app:stochastic}

Stochastic decoding preserves the main temporal structure at the prompt level.
The auxiliary stochastic suite uses Llama stress conditions: Base, DecoyA (prefix decoy), DecoyB (token decoy), and PivotC (multi-turn pivot), with 10 samples per prompt across temperature $\tau\in\{0.0,0.3,0.7,1.0\}$ and nucleus sampling $p\in\{0.9,0.95,1.0\}$.
We aggregate repeated samples per prompt before computing reliability, avoiding trial-level pseudo-replication.
DecoyB produced no LLM-judge successes in the companion AUC analysis, so the compact ICC summary below reports the three nonzero-success conditions.

\begin{table}[h]
    \centering
    \caption{Prompt-level stochastic decoding robustness for trajectory metrics. ICC values summarize stability across repeated stochastic samples; $S_0$ is deterministic for a fixed prompt and is excluded from this reliability table.}
    \label{tab:stochastic_icc}
    \small
    \begin{tabular}{lccc}
        \toprule
        Metric & Base & DecoyA & PivotC \\
        \midrule
        $t_{\text{cross}}/T$ ICC & 0.944 & 0.927 & 0.919 \\
        Sign-flip ICC & 0.899 & 0.919 & 0.925 \\
        \bottomrule
    \end{tabular}
\end{table}

\subsection{Activation Timing Sensitivity}
\label{app:tcross_sensitivity}

The first-crossing definition is not an artifact of a one-step fluctuation.
Generation trajectories can oscillate around zero after the initial transition, so we test a stricter \emph{confirmed crossing} variant that requires $n$ consecutive refusal-dominant steps:
\begin{equation}
    t_{\text{cross}}^{(n)} =
    \min\{t \geq 1 : S_t < 0,\ldots,S_{t+n-1}<0\},
\end{equation}
with right-censoring at $T$ when no confirmed crossing occurs.
The default metric is $t_{\text{cross}}^{(1)}$.
On the current 12-condition Llama-Guard grid, the $n=2$ criterion slightly increases both AUC and effect size, while $n=3$ remains significant but becomes more conservative because short or oscillatory trajectories are censored.

\begin{table}[h]
    \centering
    \caption{$t_{\text{cross}}$ sensitivity to stricter confirmed-crossing thresholds. Global rows use the 12-condition harmful grid ($n=720$); selected per-condition AUC rows show representative stable and boundary cases.}
    \label{tab:tcross_sensitivity}
    \small
    \begin{tabular}{@{}lccc@{}}
        \toprule
        & $n=1$ & $n=2$ & $n=3$ \\
        \midrule
        \multicolumn{4}{l}{\textit{Global, all conditions pooled}} \\
        \quad AUC & 0.784 & 0.791 & 0.770 \\
        \quad Cohen's $d$ & 0.796 & 0.830 & 0.440 \\
        \quad $p$-value & $<10^{-20}$ & $<10^{-20}$ & $<10^{-7}$ \\
        \midrule
        \multicolumn{4}{l}{\textit{Per-condition AUC, selected}} \\
        \quad Llama + DI & 0.799 & 0.809 & 0.836 \\
        \quad Llama + GCG & 0.987 & 0.938 & 0.925 \\
        \quad Gemma + DI & 0.792 & 0.828 & 0.862 \\
        \quad Gemma + GCG & 0.629 & 0.719 & 0.651 \\
        \quad Qwen + DI & 0.653 & 0.675 & 0.652 \\
        \quad Qwen + GCG & 0.519 & 0.502 & 0.560 \\
        \bottomrule
    \end{tabular}
\end{table}

The same qualitative pattern holds under stricter temporal criteria.
Strong in-generation conditions retain high temporal discriminability, while Qwen+GCG remains near chance, supporting the Null-generation Bias interpretation as an observability boundary rather than a threshold artifact.

\subsection{Learned Baseline Refresh}
\label{app:learned_baseline}

The trained-classifier check asks whether a fitted classifier can extract substantially more success-label information from the same $S_t$ trajectory features.
On the 12-condition Llama-Guard grid, logistic regression improves AUC modestly but does not replace the fixed temporal diagnostics.
The 5-feature set is $\{S_0,\bar{S},\bar{S}_{1{:}1},\bar{S}_{1{:}3},\bar{S}_{1{:}5}\}$; the 9-feature set adds $\min_t S_t$, $\max_t S_t$, $\mathrm{std}_t(S_t)$, and range; the 13-feature set further adds positive-step rate, $t_{\text{cross}}$, $t_{\text{cross}}/T$, and sign flips.

\begin{table}[h]
    \centering
    \caption{Learned baseline comparison on the 12-condition grid ($n=720$ harmful samples; 5-fold stratified cross-validation for LogReg rows). Fixed metrics require no training.}
    \label{tab:learned_baseline_12}
    \small
    \begin{tabular}{lccc}
        \toprule
        Method & \# Features & AUC & 95\% CI \\
        \midrule
        \multicolumn{4}{l}{\textit{Fixed metrics}} \\
        \quad $t_{\text{cross}}$ & 1 & 0.784 & [0.750, 0.816] \\
        \quad $\bar{S}_{1{:}5}$ & 1 & 0.771 & [0.738, 0.804] \\
        \quad $\bar{S}$ & 1 & 0.758 & [0.723, 0.792] \\
        \quad $S_0$ & 1 & 0.680 & [0.641, 0.718] \\
        \midrule
        \multicolumn{4}{l}{\textit{Logistic regression}} \\
        \quad LogReg($\bar{S}_{1{:}5}$) & 1 & 0.769 & [0.734, 0.802] \\
        \quad LogReg($S_0+\bar{S}$) & 2 & 0.763 & [0.728, 0.797] \\
        \quad LogReg(5 temporal) & 5 & 0.813 & [0.781, 0.844] \\
        \quad LogReg(9 trajectory) & 9 & 0.806 & [0.774, 0.837] \\
        \quad LogReg(13 features) & 13 & 0.826 & [0.795, 0.856] \\
        \bottomrule
    \end{tabular}
\end{table}

The fitted 13-feature classifier gains 4.2 pp over fixed $t_{\text{cross}}$ and 5.5 pp over fixed $\bar{S}_{1{:}5}$.
This shows that additional trajectory features contain modest complementary information, but the primary contribution remains the interpretable temporal diagnostic rather than maximum-AUC success prediction.

\subsection{Non-Lexical Baseline}
\label{app:baseline}
\label{app:nonlexical}

Non-lexical controls show that generic logit statistics do not explain away the compliance--refusal margin.
On a controlled Llama subset, entropy is informative and logit norm is near random, while the margin keeps the interpretation tied to refusal--compliance competition.

\begin{table}[h]
\centering
\caption{Non-lexical control on a controlled Llama subset ($n=60$).}
\small
\begin{tabular}{lcc}
\toprule
Metric & AUC & Feature group \\
\midrule
$\bar{S}$ & 0.932 & Logit margin \\
Compliance LogP & 0.932 & Lexicon component \\
Refusal LogP & 0.915 & Lexicon component \\
Entropy & 0.898 & Non-lexical \\
Logit Norm & 0.441 & Non-lexical \\
\bottomrule
\end{tabular}
\end{table}

\subsection{Lexicon Perturbation}
\label{app:lexicon_ablation}

The observed signal is stable under reasonable lexicon perturbations.
Because this controlled Llama subset contains few successful attacks, we treat it as a sanity check rather than primary robustness evidence.
All tested variants retain similar discrimination on this subset, indicating that the result is not driven by a single refusal or compliance token.
We test five variants on Llama-3.1-8B-Instruct ($n=60$): \textbf{original}, \textbf{no\_sorry}, \textbf{extended\_refusal}, \textbf{extended\_compliance}, and \textbf{minimal} (5-entry: \texttt{Sorry}, \texttt{illegal}, \texttt{harmful}, \texttt{Sure}, \texttt{Step 1}).

\begin{table}[h]
\centering
\caption{Lexicon perturbation results with LLM-judge success labels ($n=60$).}
\small
\begin{tabular}{lccccc}
\toprule
Variant & $S_0$ ($\mu \pm \sigma$) & $\bar{S}$ ($\mu \pm \sigma$) & AUC($S_0$) & AUC($\bar{S}$) & $\Delta$AUC \\
\midrule
original & $+1.55 \pm 1.95$ & $-1.44 \pm 0.73$ & 0.967 & 0.971 & --- \\
no\_sorry & $+2.21 \pm 1.78$ & $-1.39 \pm 0.73$ & 0.960 & 0.971 & $+0.000$ \\
extended\_refusal & $-0.43 \pm 2.07$ & $-1.85 \pm 0.67$ & 0.964 & 0.989 & $+0.018$ \\
extended\_compliance & $+4.00 \pm 1.95$ & $+0.52 \pm 0.55$ & 0.964 & 0.964 & $-0.007$ \\
minimal & $-2.61 \pm 1.74$ & $-1.57 \pm 0.49$ & 0.971 & 0.978 & $+0.007$ \\
\bottomrule
\end{tabular}
\end{table}

\begin{figure}[h]
    \centering
    \includegraphics[width=\textwidth]{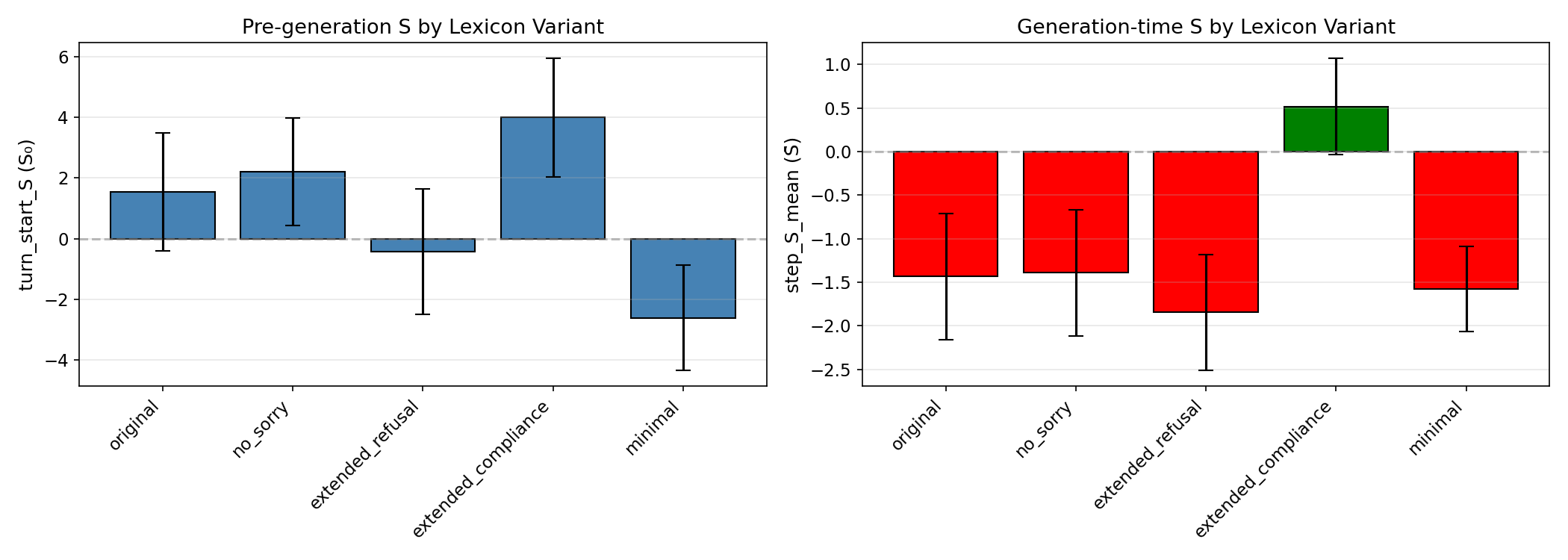}
    \caption{\textbf{Logit-Margin Score distributions by lexicon variant.}
    (Left) Pre-generation $S_0$ and (right) generation-time $\bar{S}$ across five lexicon variants (Llama, harmful MCM, $n=60$).
    Absolute scale shifts with lexicon composition, but discriminatory structure is preserved across all variants; even \texttt{extended\_compliance}, which shifts the baseline upward, maintains high AUC.}
    \label{fig:lexicon_perturbation}
\end{figure}

\paragraph{Key findings.}
On this subset, all variants achieve AUC $\geq 0.964$ (max $|\Delta\text{AUC}| = 0.018$).
Even this 5-entry minimal lexicon maintains similar discrimination (AUC $=0.978$), supporting the sanity check while leaving the primary evidence to the pooled grid.
Figure~\ref{fig:lexicon_perturbation} visualizes the score distributions across variants.

\subsection{Lexicon Aggregation Top-\texorpdfstring{$k$}{k} Sensitivity}
\label{app:topk_aggregation}

The default aggregation size $k=10$ balances two effects: smaller values are more sensitive to single-token spikes, while larger values increasingly average over lower-ranked lexicon tokens.
On the Llama+MCM harmful subset, changing $k$ alters absolute scale but preserves the refusal-dominant generation-time interpretation.

\begin{table}[h]
\centering
\caption{Effect of lexicon aggregation size $k$ on LMS summaries (Llama+MCM harmful subset; mean $\pm$ standard deviation, $n=60$).}
\label{tab:topk_aggregation}
\small
\begin{tabular}{@{}cccc@{}}
\toprule
$k$ & $S_0$ & $\bar{S}$ & Positive-step rate \\
\midrule
5  & $+0.57 \pm 2.57$ & $-1.21 \pm 0.85$ & $0.36 \pm 0.13$ \\
10 & $+1.45 \pm 2.02$ & $-1.42 \pm 0.73$ & $0.27 \pm 0.12$ \\
25 & $+1.40 \pm 1.55$ & $-1.47 \pm 0.53$ & $0.15 \pm 0.09$ \\
50 & $+0.30 \pm 1.05$ & $-1.09 \pm 0.34$ & $0.12 \pm 0.08$ \\
\bottomrule
\end{tabular}
\end{table}

\subsection{Metric Decomposition}
\label{app:metric_decomposition}

A natural concern is whether the compliance lexicon contributes genuine discriminative signal or merely reflects response-onset artifacts such as ``Sure'' and ``Certainly'' appearing at the beginning of any response.
We therefore decompose $S_t = \mu_{\text{cmp}}(t) - \mu_{\text{ref}}(t)$ into three variants:
$S_{\text{margin}}$ (the full margin used in the paper),
$S_{\text{ref}}=-\mu_{\text{ref}}$ (refusal-only),
and $S_{\text{cmp}}=\mu_{\text{cmp}}$ (compliance-only).
The margin form matters because it removes common logit movement shared by compliance and refusal lexicons.
Together with the lexicon perturbation check above, this decomposition tests whether the diagnostic is merely tracking stylistic response-openers rather than safety-relevant compliance--refusal competition.

\begin{table}[h]
\centering
\caption{\textbf{Metric decomposition: AUC comparison.} The primary 12-condition summary uses the current Llama-Guard grid ($n=720$). The detailed temporal rows report the 11-condition ablation subset used for component-wise diagnostics, where all three component metrics were available. Z-scored rows remove inter-model baseline shifts.}
\label{tab:metric_decomposition}
\small
\begin{tabular}{lcccc}
\toprule
Metric & $S_{\text{margin}}$ & $S_{\text{ref}}$ ($-\mu_{\text{ref}}$) & $S_{\text{cmp}}$ ($\mu_{\text{cmp}}$) & $\Delta_{\text{ref}-\text{margin}}$ \\
\midrule
\multicolumn{5}{l}{\textit{Primary 12-condition grid}} \\
\quad Primary temporal summary & 0.784 & 0.622 & 0.518 & $-$0.162 \\
\midrule
\multicolumn{5}{l}{\textit{Raw AUC, component-wise ablation subset}} \\
\quad $S_0$ (pre-generation)                  & 0.598 & 0.580 & 0.505 & $-$0.017 \\
\quad $\bar{S}$ (generation mean)             & 0.787 & 0.622 & 0.518 & $-$0.165 \\
\quad $\bar{S}_{1{:}5}$                       & 0.719 & 0.619 & 0.565 & $-$0.100 \\
\quad $t_{\text{cross}}$                      & 0.785 & 0.500 & ---   & $-$0.285 \\
\midrule
\multicolumn{5}{l}{\textit{Z-scored within-model AUC, component-wise ablation subset}} \\
\quad $\bar{S}$                               & 0.806 & 0.718 & 0.545 & $-$0.088 \\
\quad $\bar{S}_{1{:}5}$                       & 0.716 & 0.659 & 0.586 & $-$0.057 \\
\bottomrule
\end{tabular}
\end{table}

\paragraph{Common-mode rejection.}
The two component means are highly correlated ($r=0.979$), sharing a dominant common factor that reflects per-step logit scale rather than safety-specific evidence.
The shared component $(\mu_{\text{ref}}+\mu_{\text{cmp}})/2$ achieves only AUC $=0.594$ in the ablation subset, while the differential score reaches $0.787$.
This confirms that the margin isolates safety-relevant variation by canceling shared movement, analogous to differential signal processing.
The normalization is not merely a per-model correction: even after within-model z-scoring, the margin retains an $+0.088$ AUC advantage over refusal-only for generation-mean summaries.

\paragraph{Compliance is not merely an onset artifact.}
The margin outperforms refusal-only in most per-condition comparisons (Table~\ref{tab:decomp_percondition}), with the largest gains in conditions where compliance variation is outcome-dependent.
For example, in Gemma conditions, $\mu_{\text{cmp}}$ can differ sharply between successes and failures, so the compliance side adds signal rather than only marking generic response starts.
Refusal-only and compliance-only variants can be informative in individual conditions, but the full margin is the most stable cross-condition summary.

\begin{table}[h]
\centering
\caption{\textbf{Per-condition $\bar{S}_{1{:}5}$ AUC by metric variant.} Component-wise ablation subset. Bold indicates the best metric per condition.}
\label{tab:decomp_percondition}
\small
\begin{tabular}{lccccc}
\toprule
Condition & $n$ & ASR & $S_{\text{margin}}$ & $S_{\text{ref}}$ & $S_{\text{cmp}}$ \\
\midrule
Gemma+GCG             & 60 & 33.3\% & \textbf{0.693} & 0.511          & 0.667          \\
Gemma+MCM             & 60 & 23.3\% & \textbf{0.739} & 0.522          & 0.616          \\
Llama+DI              & 60 & 55.0\% & 0.750          & \textbf{0.767} & 0.730          \\
Llama+GCG             & 60 &  8.3\% & \textbf{0.964} & 0.935          & 0.938          \\
Llama+MCM             & 60 & 11.7\% & \textbf{0.677} & 0.663          & 0.612          \\
Mistral+DI            & 60 & 80.0\% & 0.634          & 0.526          & \textbf{0.741} \\
Mistral+GCG           & 60 & 56.7\% & \textbf{0.827} & 0.761          & 0.719          \\
Mistral+MCM           & 60 &  5.0\% & \textbf{0.708} & 0.684          & 0.643          \\
Qwen+DI               & 60 & 58.3\% & 0.627          & 0.602          & \textbf{0.639} \\
Qwen+GCG              & 60 & 76.7\% & 0.550          & 0.528          & \textbf{0.618} \\
\bottomrule
\end{tabular}
\end{table}

\subsection{Data-Driven Lexicon Induction}
\label{app:lexicon_induction}

The handcrafted lexicon used throughout this paper was designed for interpretability and cross-model portability.
A natural concern is whether results depend critically on specific lexicon choices, and whether a data-driven alternative would yield substantially different conclusions.
We address this through log-odds-based lexicon induction and cross-tokenizer analysis.

\paragraph{Cross-tokenizer analysis.}
We first verify that the handcrafted lexicon remains compact and functional across all four model tokenizers.
Table~\ref{tab:cross_tokenizer} reports the number of token IDs produced by expanding each lexicon phrase through each model's tokenizer, including bare and whitespace-prefixed forms.

\begin{table}[h]
    \centering
    \caption{Cross-tokenizer expansion of the handcrafted lexicon. The same 47 lexicon phrases (26 refusal + 21 compliance) expand to 74--102 unique token IDs depending on tokenizer granularity.}
    \label{tab:cross_tokenizer}
    \small
    \begin{tabular}{lcccc}
        \toprule
        Model & Vocab Size & Ref IDs & Cmp IDs & Lexicon \% \\
        \midrule
        Llama-3.1-8B & 128,000 & 51 & 52 & 0.080\% \\
        Qwen2.5-7B   & 151,643 & 51 & 52 & 0.067\% \\
        Mistral-7B   &  32,768 & 36 & 40 & 0.226\% \\
        Gemma-2-9B   & 256,000 & 48 & 52 & 0.038\% \\
        \bottomrule
    \end{tabular}
\end{table}

The lexicon occupies less than 0.23\% of each vocabulary, yet achieves consistent AUC across all four models.
Mistral's smaller vocabulary produces fewer token IDs due to coarser BPE segmentation, but this does not degrade performance.

\paragraph{Induction method.}
We induce data-driven lexicons via word-level log-odds ratios.
Using LLM-as-judge labels (\texttt{is\_success\_llm}) from Llama-Guard-3-8B, we partition response texts into compliant and refusal-like outputs.
The compliant set contains harmful prompts where the attack succeeded; the refusal set contains failed harmful prompts plus benign responses.
For each word $w$ appearing at least five times, we compute
\[
\text{LOR}(w)=\log P(w\mid \text{compliant})-\log P(w\mid \text{refusal})
\]
with Laplace smoothing ($\alpha=1$).
The top-25 words by $|\text{LOR}|$ form the induced refusal and compliance lexicons.

\paragraph{Semantic direction.}
The induced lexicons capture semantically meaningful but more content-specific patterns.
Induced refusal words include benign-topic words such as \emph{energy}, \emph{sleep}, \emph{cells}, \emph{immune}, \emph{ingredients}, \emph{memory}, \emph{cooking}, and \emph{solar}, as well as safety markers such as \emph{cannot} and \emph{harmful}.
Induced compliance words include domain-specific harmful-content terms such as \emph{phishing}, \emph{vpn}, \emph{fetish}, \emph{stealth}, \emph{evade}, \emph{rootkit}, \emph{deepfake}, and \emph{scam}.
These differ qualitatively from the handcrafted lexicon, which targets generic safety behavior (\emph{Sorry}, \emph{cannot}, \emph{Sure}, \emph{Step~1}).
The Jaccard overlap between induced and handcrafted word sets is near zero (refusal: 0.039, compliance: 0.000 for Llama-only; 0.000/0.000 for the full dataset), confirming that the two approaches capture largely complementary signals.

\paragraph{AUC comparison.}
Table~\ref{tab:lexicon_induction_auc} compares three approaches: real-time logit-based $S_t$ metrics, a bag-of-words (BoW) proxy using the handcrafted lexicon, and a BoW proxy using the induced lexicon.
The BoW proxy computes $S_{\text{BoW}}=(n_{\text{refusal}}-n_{\text{compliance}})/(n_{\text{refusal}}+n_{\text{compliance}})$ from word counts in the full response text.
This diagnostic pool includes the primary grid plus auxiliary non-stochastic generations used for lexicon checks, so its sample sizes differ from the 12-condition primary grid.

\begin{table}[h]
    \centering
    \caption{AUC comparison: logit-based metrics vs.\ BoW lexicon proxies (label = \texttt{is\_success\_llm}; stochastic samples excluded).}
    \label{tab:lexicon_induction_auc}
    \small
    \begin{tabular}{lcc}
        \toprule
        Method & All Models ($n=2{,}040$) & Llama-only ($n=840$) \\
        \midrule
        $S_0$ (logit, real-time) & 0.630 & 0.702 \\
        $\bar{S}_{1{:}5}$ (logit, real-time) & 0.701 & 0.735 \\
        $\bar{S}$ (logit, real-time) & 0.691 & 0.787 \\
        \addlinespace
        BoW --- handcrafted lexicon & 0.674 & 0.774 \\
        BoW --- induced lexicon & \textbf{0.996} & \textbf{0.936} \\
        \bottomrule
    \end{tabular}
\end{table}

\paragraph{Interpretation and caveats.}
The induced BoW lexicon achieves substantially higher AUC (0.936--0.996) than both the handcrafted BoW proxy (0.674--0.774) and the logit-based metrics (0.630--0.787).
However, this comparison is not apples-to-apples: BoW operates on the complete response text after generation, while $S_t$ is computed online from token-level logits.
The induced BoW AUC should therefore be interpreted as an upper bound on lexicon quality, not as a direct competitor to real-time logit-level observability.
Two conclusions follow.
First, the handcrafted lexicon is a conservative lower bound: a data-driven lexicon can capture more success-label information, so the reported LMS AUC values do not exhaust the information available in token distributions.
Second, the near-zero Jaccard overlap shows that handcrafted and induced lexicons capture complementary signals: generic refusal/compliance behavior versus domain-specific harmful content.

\paragraph{Per-attack induced lexicons.}
Table~\ref{tab:per_attack_induced} shows how induced compliance words differ by attack paradigm, confirming that each attack elicits distinct harmful-content domains.

\begin{table}[h]
    \centering
    \caption{Top induced compliance words and AUC by attack paradigm (BoW-induced vs.\ BoW-handcrafted).}
    \label{tab:per_attack_induced}
    \small
    \begin{tabular}{llccc}
        \toprule
        Attack & Top Induced Compliance Words & BoW-HC & BoW-Ind & ASR \\
        \midrule
        MCM & \emph{stealth, sedative, narrative, personnel} & 0.730 & 0.977 & 10.2\% \\
        GCG & \emph{blade, credentials, posts, anal} & 0.537 & 0.969 & 46.7\% \\
        DI & \emph{exploiting, injection, scam, deepfake} & 0.500 & 0.971 & 62.8\% \\
        \bottomrule
    \end{tabular}
\end{table}

Notably, the handcrafted BoW proxy is near random for DI (0.500), indicating that DI-generated responses often avoid standard refusal keywords.
The induced lexicon attains strong discrimination (0.971) by capturing domain-specific content.
This per-attack variation reinforces the design rationale for the handcrafted lexicon: it sacrifices per-condition optimality for cross-condition consistency.

\paragraph{Transferability analysis.}
A key concern with data-driven lexicons is whether they generalize beyond the training distribution.
We evaluate cross-attack and cross-model transferability by training the induced lexicon on one slice and testing it on others.

\begin{table}[h]
    \centering
    \caption{Cross-attack transfer of induced lexicons ($n=660$, all models). Diagonal entries (*) are in-distribution; off-diagonal entries are out-of-distribution.}
    \label{tab:cross_attack_transfer}
    \small
    \begin{tabular}{lccc}
        \toprule
        Train $\to$ Test & DI & GCG & MCM \\
        \midrule
        DI & 0.891* & 0.670 & 0.530 \\
        GCG           & 0.688  & 0.932* & 0.676 \\
        MCM           & 0.606  & 0.734 & 0.890* \\
        \bottomrule
    \end{tabular}
\end{table}

\begin{table}[h]
    \centering
    \caption{Cross-model transfer of induced lexicons ($n=660$, all attacks). Diagonal entries (*) are in-distribution.}
    \label{tab:cross_model_transfer}
    \small
    \begin{tabular}{lcccc}
        \toprule
        Train $\to$ Test & Gemma & Llama & Mistral & Qwen \\
        \midrule
        Gemma   & 0.970* & 0.817 & 0.662 & 0.544 \\
        Llama   & 0.673  & 1.000* & 0.705 & 0.682 \\
        Mistral & 0.594  & 0.729 & 0.968* & 0.692 \\
        Qwen    & 0.615  & 0.522 & 0.642 & 0.957* \\
        \bottomrule
    \end{tabular}
\end{table}

Induced lexicons transfer moderately across conditions (mean off-diagonal AUC $\approx 0.65$) but substantially underperform their in-distribution AUC ($\approx 0.93$).
This confirms that induced lexicons capture condition-specific vocabulary rather than universal safety behavior.
The refusal word overlap across attack-specific lexicons is modest (Jaccard $=0.11$--$0.25$ for refusal words; $0.02$--$0.04$ for compliance words), explaining the transfer gap.
Common refusal words across all conditions are often benign topic indicators rather than safety markers, while compliance words are almost entirely condition-specific.

\paragraph{Compatibility with temporal decomposition.}
The paper's central diagnostic relies on the temporal structure of the $S_t$ trajectory, specifically the decomposition into pre-generation ($S_0$) and generation-time ($\bar{S}$) components.
This decomposition is fundamentally unavailable to post-hoc text classifiers, including induced BoW proxies.
The induced BoW score computes a single scalar from the complete response text, collapsing all temporal information into one value.
Consequently, $\text{RP}_A$ cannot be computed because it requires $S_0$, $\text{RP}_B$ cannot be computed because it requires comparing $S_0$ against $\bar{S}$, sign reversal cannot be measured as a pre-generation-to-generation transition, and early-$k$ detection curves cannot be evaluated without per-token logit access.
In short, induced lexicons support the information content of safety-relevant vocabulary, but they cannot replace the temporal decomposition that makes LMS informative.
The two approaches are complementary: logit-based $S_t$ reveals when and how safety signals emerge, while data-driven lexicon induction confirms that vocabulary choice is not the limiting source of signal.

\paragraph{Surface-token and tokenizer audit scope.}
A remaining concern is that LMS may partially reflect surface-token or tokenizer artifacts rather than only a distributed compliance--refusal contrast.
This concern is most relevant for multi-token surface phrases whose tokenizer expansion can introduce frequent subword pieces or digit tokens.
Our lexicon perturbation, metric-decomposition, and data-driven lexicon checks reduce this concern, but they do not eliminate the need for finer-grained tokenizer audits.
Future audits should report token-contribution concentration, leave-one-token-out sensitivity, digit-token removal, whitespace-prefixed-token variants, and response-onset-token removal to test whether LMS is dominated by a small number of tokenizer-specific fragments.

\subsection{Probe-Derived Lexicon Check}
\label{app:probe_lexicon}

Probe-derived vocabulary rankings confirm that many handcrafted tokens align with model-internal refusal/compliance directions.
We project vocabulary embeddings onto a refusal direction and report representative rank percentiles for single-token lexicon entries.

\begin{table}[h]
    \centering
    \caption{Probe-derived rank percentiles for representative handcrafted tokens. Higher values indicate stronger alignment with the intended refusal or compliance side.}
    \label{tab:probe_refusal}
    \small
    \begin{tabular}{lcccc}
        \toprule
        Token & Llama & Mistral & Qwen & Gemma \\
        \midrule
        \multicolumn{5}{l}{\textit{Refusal tokens, percentile from top}} \\
        \texttt{illegal} & 85.7\% & 100.0\% & 65.9\% & 99.4\% \\
        \texttt{harmful} & 91.4\% & 98.1\% & 97.7\% & 99.2\% \\
        \texttt{dangerous} & 64.9\% & 100.0\% & 77.4\% & 99.2\% \\
        \texttt{Sorry} & 89.0\% & 89.4\% & 60.8\% & 50.3\% \\
        \midrule
        \multicolumn{5}{l}{\textit{Compliance tokens, percentile from bottom}} \\
        \texttt{Sure} & 90.0\% & 37.5\% & 51.1\% & 46.9\% \\
        \texttt{Step} & 90.3\% & 43.6\% & 95.5\% & 78.6\% \\
        \texttt{Below} & 50.6\% & 98.9\% & 93.0\% & 96.4\% \\
        \texttt{Procedure} & 96.6\% & 49.3\% & 70.6\% & 33.5\% \\
        \bottomrule
    \end{tabular}
\end{table}

The ranking is not perfect, especially for compliance tokens, but the direction is consistent enough to support the handcrafted lexicon as an interpretable lower-access choice.

\subsection{Vocabulary Top-\texorpdfstring{$k$}{k} Logit Access}
\label{appendix:topk_truncation}

The diagnostic is hidden-state-free, but it is not fully black-box under APIs that expose only a small top-$k$ set of output log-probabilities.
A key practical question is whether $S_t$ can be computed from the limited log-probability information returned by closed-source APIs.
For example, an API may expose only a small \texttt{top\_logprobs} window rather than the full vocabulary distribution.
To test this directly, we simulate vocabulary-level top-$k$ truncation on Llama-3.1-8B-Instruct across MCM, GCG, and DI (60 samples per attack).

\paragraph{Method.}
For each decoding step $t$, we take the full logit vector $\ell_t \in \mathbb{R}^{|\mathcal{V}|}$ and construct a truncated vector $\hat{\ell}^{(k)}_t$ by retaining only the $k$ highest-valued entries and setting all others to $-\infty$.
We then recompute $S_t$ from $\hat{\ell}^{(k)}_t$, counting only lexicon tokens that survive in the top-$k$ window.
We evaluate $k\in\{5,10,20,50,100,200\}$ and measure Pearson correlation with the full-logit $S_t$ trajectory, mean absolute error (MAE), sign agreement, and refusal/compliance lexicon survival rates.
This is distinct from Appendix~\ref{app:topk_aggregation}: lexicon-internal aggregation assumes the lexicon token logits are available, whereas vocabulary-level truncation asks whether those safety-relevant tokens appear in the API-visible top-$k$ window at all.

\begin{table}[h]
    \centering
    \caption{Vocabulary-level top-$k$ truncation simulation (Llama-3.1-8B-Instruct, $n=60$ per attack).
    Pearson $r$ is the correlation between full-logit and truncated $S_t$ trajectories.
    Sign Agr. is the fraction of steps preserving $S_t$ polarity.
    Ref/Cmp Surv. is the mean fraction of refusal/compliance lexicon tokens surviving in the top-$k$ window.}
    \label{tab:topk_truncation}
    \small
    \begin{tabular}{llccccc}
        \toprule
        Attack & $k$ & Pearson $r$ & MAE & Sign Agr. & Ref Surv. & Cmp Surv. \\
        \midrule
        MCM & 5 & $-$0.199 & 5.023 & 0.637 & 1.2\% & 0.6\% \\
         & 10 & $-$0.108 & 4.779 & 0.661 & 1.9\% & 1.0\% \\
         & 20 & $-$0.120 & 4.254 & 0.576 & 2.7\% & 1.9\% \\
         & 50 & 0.111 & 2.899 & 0.576 & 5.1\% & 3.8\% \\
         & 100 & 0.319 & 2.503 & 0.586 & 7.5\% & 5.4\% \\
         & 200 & 0.472 & 2.236 & 0.619 & 10.6\% & 7.5\% \\
        \addlinespace
        GCG & 5 & $-$0.106 & 4.489 & 0.477 & 1.5\% & 0.8\% \\
         & 10 & $-$0.071 & 4.442 & 0.520 & 2.3\% & 1.5\% \\
         & 20 & $-$0.009 & 3.817 & 0.564 & 3.2\% & 2.6\% \\
         & 50 & $-$0.059 & 3.406 & 0.538 & 5.2\% & 4.3\% \\
         & 100 & $-$0.051 & 3.108 & 0.541 & 7.6\% & 6.3\% \\
         & 200 & 0.042 & 2.674 & 0.588 & 10.9\% & 8.5\% \\
        \addlinespace
        DI & 5 & 0.121 & 5.638 & 0.470 & 0.7\% & 0.5\% \\
         & 10 & $-$0.089 & 7.107 & 0.445 & 1.5\% & 0.8\% \\
         & 20 & $-$0.072 & 6.022 & 0.523 & 2.0\% & 1.5\% \\
         & 50 & 0.117 & 4.651 & 0.612 & 2.9\% & 3.1\% \\
         & 100 & 0.231 & 3.561 & 0.670 & 4.3\% & 4.6\% \\
         & 200 & 0.387 & 2.773 & 0.673 & 6.5\% & 6.5\% \\
        \bottomrule
    \end{tabular}
\end{table}

\paragraph{Results.}
The results reveal a large gap between lexicon-internal aggregation stability and vocabulary-level truncation.
At $k=20$, a typical API-like level, only 2--3\% of refusal lexicon tokens and 1.5--2.6\% of compliance tokens survive in the top-$k$ window.
Trajectory correlation with the full-logit $S_t$ trajectory is near zero or negative ($r=-0.120$ for MCM, $r=-0.009$ for GCG, and $r=-0.072$ for DI), and sign agreement is only 0.523--0.576.
Even at $k=200$, GCG remains near zero ($r=0.042$), consistent with adversarial suffix optimization dispersing probability mass across unusual token positions that rarely appear in the top-$k$ window.

\paragraph{Implications.}
This simulation confirms that the current formulation requires full logits or targeted access to the relevant lexicon token log-probabilities.
Standard top-$k$ log-probability outputs are insufficient for reliable LMS computation.
This empirically supports positioning $S_t$ as a diagnostic for open-weight model auditing rather than a universal deployment-time detector, and motivates future work on API-compatible reformulations such as targeted logit-bias probing or lightweight distilled monitors.

\subsection{Compliance-Dominant Successful Jailbreaks}
\label{app:compliance_dominant}

Successful harmful generations are usually compliance-dominant in the aggregate signal, which is why LMS should be read temporally rather than as a one-shot polarity classifier.
In the primary 12-condition grid, 92.3\% of successful jailbreaks (263/285) have $\bar{S}>0$, compared with 49.7\% of failed jailbreaks (216/435).
Table~\ref{tab:compliance_dominant} reports the successful-jailbreak breakdown by condition.

\begin{table}[h]
    \centering
    \caption{Successful jailbreaks with compliance-dominant generation ($\bar{S}>0$) in the 12-condition grid.}
    \label{tab:compliance_dominant}
    \small
    \begin{tabular}{@{}llrrr@{}}
        \toprule
        Model & Attack & $n_{\text{succ}}$ & $n_{\bar{S}>0}$ & Rate \\
        \midrule
        Gemma & DI & 23 & 23 & 100.0\% \\
        Gemma & GCG & 20 & 20 & 100.0\% \\
        Gemma & MCM & 11 & 11 & 100.0\% \\
        Llama & DI & 33 & 30 & 90.9\% \\
        Llama & GCG & 5 & 5 & 100.0\% \\
        Llama & MCM & 3 & 3 & 100.0\% \\
        Mistral & DI & 48 & 48 & 100.0\% \\
        Mistral & GCG & 34 & 31 & 91.2\% \\
        Mistral & MCM & 3 & 3 & 100.0\% \\
        Qwen & DI & 35 & 35 & 100.0\% \\
        Qwen & GCG & 46 & 42 & 91.3\% \\
        Qwen & MCM & 24 & 12 & 50.0\% \\
        \midrule
        \textbf{All} & & \textbf{285} & \textbf{263} & \textbf{92.3\%} \\
        \bottomrule
    \end{tabular}
\end{table}

\section{Judge and Behavioral Probe Checks}
\label{app:judge}

\subsection{Judge Scope}

The authoritative 12-condition ledger in Appendix~\ref{app:authoritative_12} uses Llama-Guard-3-8B as the primary judge.
HarmBench-Mistral-7B and GPT-4o have also been refreshed on the same 12-condition grid.
The auxiliary judges support the sample-level prediction results but show condition-level label disagreements, so the main evidence should be read as primary-judge evidence rather than a full judge-invariance claim.

\begin{table}[h]
    \centering
    \caption{Primary and refreshed auxiliary judge summary on the same 12-condition grid. Success-label agreement is harmful-sample agreement with Llama-Guard. PGB/IGB/NGB annotation match counts condition-level agreement after the same permutation rule, excluding auxiliary conditions with insufficient successful samples.}
    \label{tab:judge_refresh}
    \small
    \resizebox{\textwidth}{!}{%
    \begin{tabular}{lccccc}
        \toprule
        Judge & ASR & Success-label agreement & $\bar{S}_{1{:}5}$ AUC & $t_{\text{cross}}$ AUC & Annotation match \\
        \midrule
        Llama-Guard & 39.6\% & -- & 0.771 & 0.784 & 12/12 \\
        HarmBench & 33.3\% & 84.3\% & 0.752 & 0.734 & 10/12 \\
        GPT-4o & 42.9\% & 87.8\% & 0.820 & 0.804 & 8/11$^\dagger$ \\
        \bottomrule
    \end{tabular}
    }
    \\[2pt]
    \footnotesize{$^\dagger$Llama+MCM has only two GPT-4o successes and is reported as N/A for auxiliary RP-plane annotation.}
\end{table}

HarmBench disagrees with the primary RP-plane annotations on Mistral+MCM and Qwen+MCM, both cases where refreshed judge labels yield axis-dominant PGB while the primary Llama-Guard permutation stage assigns NGB.
GPT-4o agrees on 8 of 11 available annotations; its disagreements are Mistral+MCM and Qwen+MCM as PGB, plus Qwen+DI as IGB, while the primary judge assigns all three to NGB.

\subsection{Adjudicated Two-Rater Human Audit}
\label{app:human_audit}

We additionally constructed a 240-item adjudicated human reference audit using two internal author-raters.
The two raters independently labeled each response as safe/refusal, ambiguous/partial, or harmful compliance; all disagreements were then reconciled into a final adjudicated label.
The audit intentionally oversamples automatic-judge disagreements and balanced success/failure cases, so its label proportions are not ASR estimates.
Instead, the audit measures label reliability and compares binary judges against a human-adjudicated reference subset.

\begin{table}[h]
    \centering
    \caption{Adjudicated two-rater human audit of response-level jailbreak-success labels. Human labels use three categories: safe/refusal, ambiguous/partial, and harmful compliance. Binary harmful-compliance agreement collapses safe/refusal and ambiguous/partial into the non-compliance side.}
    \label{tab:human_audit}
    \small
    \begin{tabular}{lc}
        \toprule
        Quantity & Value \\
        \midrule
        Items labeled by both raters & 240 \\
        Exact three-class agreement & 193/240 (80.4\%) \\
        Cohen's $\kappa$ (three-class) & 0.698 \\
        Binary harmful-compliance agreement & 213/240 (88.8\%) \\
        Cohen's $\kappa$ (binary harmful-compliance) & 0.732 \\
        Disagreements reconciled & 47/47 (100.0\%) \\
        Final adjudicated labels & 74 safe / 108 ambiguous / 58 harmful \\
        \bottomrule
    \end{tabular}
\end{table}

After adjudication, 132 items have non-ambiguous safe/refusal or harmful-compliance labels and can be directly compared with binary judges.
On this adjudicated non-ambiguous subset, GPT-4o matched the human label on 103/132 items, Llama-Guard on 102/132 items, and HarmBench on 88/132 items.
Thus the human audit supports the paper's judge-conditioned framing: LLM-as-judge labels are useful response-level outcome variables, but borderline partial-compliance outputs remain a material source of label ambiguity rather than a solved reference-label problem.

\subsection{Primary-Judge Intervention Summary}
\label{app:intervention_judges}

The early-stopping procedure is used as a behavioral probe, not as a proposed complete defense.
The primary 12-condition Llama-Guard run is reported in Table~\ref{tab:type_ab}: aggregate ASR decreases from 39.6\% to 13.1\% under the $w=5$ probe.
Because the behavioral probe is defined from $t_{\text{cross}}$, it is not independent of the temporal diagnostic.

\begin{table}[h]
    \centering
    \caption{Early-stopping behavioral probe response by coarse RP-plane annotation. The response pattern shows behaviorally distinct responses, not an independent source of evidence; format-free benign FPR reports the separate natural-benign query check.}
    \label{tab:intervention}
    \label{tab:intervention_by_type_judges}
    \small
    \setlength{\tabcolsep}{4.0pt}
    \begin{tabular}{lccccc}
        \toprule
        Annotation & Original ASR & Probe ASR & $\Delta$ASR & \shortstack{Format-free\\benign FPR} & Reading \\
        \midrule
        PGB & 33.3\% & 19.4\% & 13.9 pp & 0.0\% & Small/selective effect \\
        IGB & 28.8\% & 10.0\% & 18.8 pp & 0.0\% & Clear generation-time effect \\
        NGB & 52.0\% & 11.7\% & 40.3 pp & 0.0\% & Large but least diagnostic/selective \\
        \bottomrule
    \end{tabular}
\end{table}

\subsection{\texorpdfstring{$S_0$-Gated Stress Test}{S0-Gated Stress Test}}
\label{app:s0_gating}

The $S_0$-gated variant is a stress test for attack-formatted benign references in the same 12-condition grid.
Format-free benign queries have 0.0\% format-free benign FPR in a separate check; attack-formatted benign references are harder because they intentionally share the attack wrapper.
Thus, the table below should be read as wrapper-confound stress behavior rather than as the format-free benign FPR.
Formally, the original halt rule is $H_w=\mathbf{1}[t_{\text{cross}}>w]$, and the gated rule is $H_{w,\tau}^{S_0}=\mathbf{1}[t_{\text{cross}}>w \wedge S_0<\tau]$.

\begin{table}[h]
    \centering
    \caption{$S_0$-gated vs.\ original early stopping on attack-formatted benign reference stress tests in the 12-condition grid ($w=5$, $\tau=6.3$).}
    \label{tab:s0_gating}
    \small
    \begin{tabular}{@{}lcccc@{}}
        \toprule
        & \multicolumn{2}{c}{Original} & \multicolumn{2}{c}{$S_0$-Gated} \\
        \cmidrule(lr){2-3} \cmidrule(lr){4-5}
        & $\Delta$ASR & \shortstack{Attack-formatted\\false halt} & $\Delta$ASR & \shortstack{Attack-formatted\\false halt} \\
        \midrule
        Aggregate & 26.5 pp & 46.0\% & 17.1 pp & 12.6\% \\
        Gemma   & 20.0 pp & 37.8\% & 10.6 pp &  0.0\% \\
        Llama   & 17.8 pp & 52.8\% & 14.4 pp &  6.1\% \\
        Mistral & 27.8 pp & 45.6\% & 27.8 pp & 44.4\% \\
        Qwen    & 40.6 pp & 47.8\% & 16.1 pp &  0.0\% \\
        \bottomrule
    \end{tabular}
\end{table}

\section{Reproducibility Resources}
\label{app:reproducibility_resources}

\subsection{Supplementary Archive}
\label{app:supplementary_archive}

The supplementary archive provides benchmark input prompts, prompt-construction files, LMS extraction code, reproduction scripts, aggregate result files, and redacted per-sample metric tables for the main numerical claims.
The prompt files support rerunning LMS extraction from model logits, while the redacted sample tables remove prompt text, generated responses, raw judge outputs, and local experiment folder names while preserving the metrics and labels needed to recompute the reported AUC, temporal, RP, hidden-alignment, and early-stopping summaries.
The archive also includes a README with reproduction commands, lightweight and generation-specific Python requirements, and an external-asset license note.

\subsection{Compute Resources}
\label{app:compute_resources}

All reported main-grid generations and hidden-state captures were run on a single local workstation.
The machine used one NVIDIA GeForce RTX 4090 GPU with 24 GB VRAM (CUDA 13.1, driver 590.48), an AMD Ryzen 9 3950X CPU with 16 cores and 32 threads, 62 GB system RAM, and a 457 GB root disk.
Idle post-experiment readings were approximately 1.4 GB GPU memory in use and 60$^\circ$C GPU temperature.

\begin{table}[h]
    \centering
    \caption{Approximate runtime for the main per-model evaluation runs, estimated from output file timestamps.}
    \label{tab:compute_runtime}
    \small
    \begin{tabular}{lcc}
        \toprule
        Run & Completion time & Approximate runtime \\
        \midrule
        Llama-3.1-8B-Instruct & 22:50 & 12 min \\
        Qwen2.5-7B-Instruct & 23:02 & 12 min \\
        Mistral-7B-Instruct-v0.3 & 23:14 & 12 min \\
        Gemma-2-9B-Instruct & 23:36 & 22 min \\
        Analysis scripts & 23:36 & $<1$ min \\
        \midrule
        Total & -- & 58 min \\
        \bottomrule
    \end{tabular}
\end{table}

Gemma took longer primarily because Gemma-2-9B has more transformer layers (42) and therefore larger hidden-state outputs.
Using the observed experiment-time GPU power draw of approximately 239 W, the total run consumed roughly $239\mathrm{W}\times3{,}480\mathrm{s}\approx832$ kJ, or 0.23 kWh.
The main capture covered four models, 240 generations per model, about 64 decoding steps per generation, and three retained analysis layers, yielding approximately 184{,}320 recorded layer-step measurements in addition to the standard logit outputs.

\subsection{External Assets and Licenses}
\label{app:asset_licenses}

Table~\ref{tab:asset_licenses} lists the external assets used in the experiments and the license or access terms reported by their official model cards or repositories.
We use these assets for evaluation only and do not redistribute third-party model weights or service outputs in this submission.
The supplementary archive includes benchmark input prompts for reproducibility under the license terms listed in Table~\ref{tab:asset_licenses}.

\begin{table}[h]
    \centering
    \caption{External assets used in the experiments.}
    \label{tab:asset_licenses}
    \scriptsize
    \setlength{\tabcolsep}{3.0pt}
    \renewcommand{\arraystretch}{1.12}
    \begin{tabularx}{\textwidth}{@{}p{0.29\textwidth}p{0.31\textwidth}X@{}}
        \toprule
        Asset & Use in this paper & License or terms \\
        \midrule
        Meta-Llama-3.1-8B-Instruct & Target model & Llama 3.1 Community License and acceptable-use terms \\
        Llama-Guard-3-8B & Primary LLM-as-judge safety classifier & Llama 3.1 Community License and acceptable-use terms \\
        Mistral-7B-Instruct-v0.3 & Target model & Apache License 2.0 \\
        Qwen2.5-7B-Instruct & Target model & Apache License 2.0 \\
        Gemma-2-9B-Instruct & Target model & Gemma license / Gemma Terms of Use \\
        JailbreakBench JBB-Behaviors & Harmful prompt source & MIT License \\
        HarmBench-Mistral-7B validation classifier & Auxiliary LLM-as-judge safety classifier & MIT License \\
        GPT-4o & Auxiliary judge refresh & OpenAI service access terms; no weights redistributed \\
        \bottomrule
    \end{tabularx}
\end{table}

\clearpage

\end{document}